\newif\ifarXiv         
\newif\ifjournal        
  \journalfalse       \arXivtrue      

 \ifarXiv   
\documentclass[11pt]{article}
\title{Identifiability of interaction kernels in \\ mean-field equations of interacting particles }
 \author{Quanjun Lang$^*$, Fei Lu\footnote{ Department of Mathematics, Johns Hopkins University. Email: qlang1@math.jhu.edu; feilu@math.jhu.edu }} 
\date{}
\fi

\usepackage{amsmath,bm,mathtools,amsfonts,mathrsfs,amssymb,amsthm,mathabx,setspace}              
\usepackage{subfigure,graphicx,color,epstopdf} 
\usepackage{float,verbatim,tabularx} 
\usepackage{times}
\usepackage[OT1]{fontenc}
\usepackage{enumerate} 
\usepackage{bbm}
\usepackage[bookmarks=true,         bookmarksnumbered=true,
colorlinks=true,
pdfstartview=FitV,
linkcolor=blue, citecolor=blue, urlcolor=blue,
]{hyperref}
\usepackage[dvipsnames]{xcolor}

\usepackage{booktabs}  

\usepackage{algorithm}
\usepackage[noend]{algpseudocode}
\usepackage[normalem]{ulem}
\usepackage[font=small]{caption}
\newcommand{\myparagraph}[1]{\vspace{1mm}\noindent\textbf{#1}}
\newtheorem{theorem}{Theorem}[section]
\newtheorem{assumption}[theorem]{Assumption}
\newtheorem{remark}[theorem]{Remark}
\newtheorem{definition}[theorem]{Definition}

\newtheorem{lemma}[theorem]{Lemma}

\newtheorem{example}[theorem]{Example}
\numberwithin{equation}{section}

\newcommand{\wbar}{\widebar}
\newcommand{\rhoT}{\wbar \rho_T}
\newcommand{\rhoTd}{\wbar \rho_T}

\newcommand{\GbarT}{{\overline G_T}}
\newcommand{\FbarT}{{\overline F_T}}
\newcommand{\QbarT}{{\overline Q_T}}
\newcommand{\RbarT}{{\overline R_T}}

\newcommand{\DrhoT}{\widehat{\overline \rho}_T}

\newcommand{\vertiii}[1]{{\left\vert\kern-0.25ex\left\vert\kern-0.25ex\left\vert #1 
    \right\vert\kern-0.25ex\right\vert\kern-0.25ex\right\vert}}

\newcommand{\barX}{\wbar X}

\newcommand{\divg}{\mathrm{div}}
\newcommand{\Ktrue}{K_{true}}
\newcommand{\mathspan}{\mathrm{span}}

\newcommand{\mbf}[1]{\boldsymbol{#1}}

\newcommand{\innerp}[2]{ #1\cdot #2 }
\newcommand{\inp}[1]{\langle{#1}\rangle}

\newcommand{\abs}[1]{\big| #1 \big|}

\newcommand{\realR}[1]{\mathbb{R}^{#1}}
\newcommand{\real}{\mathbb{R}}

\newcommand{\bu}{\mbf{u}}

\newcommand{\bB}{\mbf{B}}

\newcommand{\bX}{\mbf{X}}

\newcommand{\mE}{\mathcal{E}}

\newcommand{\mL}{\mathcal{L}}
\newcommand{\mH}{\mathcal{H}}

\newcommand{\mN}{\mathcal{N}}

\newcommand{\mX}{\mathcal{X}}

\newcommand{\R}{\real}

\newcommand{\hypspace}{\mathcal{H}}
\newcommand{\E}{\mathbb{E}}

\newcommand{\grad}[1]{\nabla #1}

\newcommand{\argmin}[1]{\underset{#1}{\operatorname{arg}\operatorname{min}}\;}

\newcommand{\supp}[1]{\text{supp}(#1)}

\newcommand{\ang}[1]{\left<#1\right>}       
\newcommand{\crl}[1]{\left\{#1\right\}}     
\newcommand{\edg}[1]{\left[#1\right]}       
\newcommand{\rkhs}[1]{\langle\hspace{-1mm} \langle #1 \rangle\hspace{-1mm} \rangle}    
\newcommand{\normmm}[1]{{\left\vert\kern-0.25ex\left\vert\kern-0.25ex\left\vert #1 
    \right\vert\kern-0.25ex\right\vert\kern-0.25ex\right\vert}}
\newcommand{\longinprod}[1]{\left< \kern-0.25ex\left< \kern-0.25ex\left<#1 \right> \kern-0.25ex\right> \kern-0.25ex\right>}    

\DeclareMathAlphabet{\mathpzc}{OT1}{pzc}{m}{it}

\usepackage{enumitem}
\usepackage{xcolor}

  \usepackage[margin=0.98in]{geometry}  

\begin{document} 

\maketitle \vspace{-6mm}

 \begin{abstract}
    \vspace{-2mm}
 This study examines the identifiability of interaction kernels in mean-field equations of interacting particles or agents, an area of growing interest across various scientific and engineering fields. The main focus is identifying data-dependent function spaces where a quadratic loss functional possesses a unique minimizer. We consider two data-adaptive $L^2$ spaces: one weighted by a data-adaptive measure and the other using the Lebesgue measure. In each $L^2$ space, we show that the function space of identifiability is the closure of the RKHS associated with the integral operator of inversion. 
 Alongside prior research, our study completes a full characterization of identifiability in interacting particle systems with either finite or infinite particles, highlighting critical differences between these two settings. Moreover, the identifiability analysis has important implications for computational practice. It shows that the inverse problem is ill-posed, necessitating regularization. Our numerical demonstrations show that the weighted $L^2$ space is preferable over the unweighted $L^2$ space, as it yields more accurate regularized estimators.    
\end{abstract} 
     \textbf{Keywords:}  mean-field equations, identifiability,  RKHS, regularization, inverse problem. 
\ifarXiv 
\fi


\vspace{-2mm}\section{Introduction}

Systems of interacting particles or agents have become increasingly used in many areas of science and engineering (see \cite{bell2005_ParticlebasedSimulation,VZ2012,MT2014,baumgarten2019_GeneralConstitutive} and the references therein). Driven by these applications, there is a growing interest in inferring the interaction kernel (or the interaction potential) from data, either parametrically \cite{kasonga1990_MaximumLikelihood,chen2021_MaximumLikelihood,sharrock2021parameter,MaestraHoffman22} or in a nonparametrically for broader applicability  \cite{BFHM17,LZTM19,LMT21_JMLR,LMT21,LangLu22,della2022nonparametric,yao2022mean}.

The inference problem can be classified into two categories: (i) a statistical learning problem when the system consists of finitely many particles and the data includes multiple trajectories of all particles, and (ii) a deterministic inverse problem for the mean-field equation (MFE) from data consisting of a solution to the MFE, which arises when the number of particles is so large that only the macroscopic density of particles can be observed. For the statistical learning of kernels in systems with finitely many particles, previous studies \cite{LZTM19,LMT21_JMLR,LMT21} minimize loss functionals based on the mean-square error or the likelihood of the data, establishing computationally efficient algorithms that yield nonparametric estimators achieving the minimax rate of convergence. In particular, the studies \cite{LLMTZ21,LiLu20} show that any square-integrable kernel is identifiable under a coercivity condition, which imposes constraints on the distribution of the data trajectories. For the inverse problem of the MFE, the study \cite{LangLu22} has introduced a
\emph{derivative-free probabilistic loss functional} and based on it, a scalable nonparametric regression algorithm that produces a convergent estimator robust to discrete noisy data. However, it remains open to study the identifiability of the kernel in the MFE from data. 

This study provides a complete characterization of the identifiability of kernels in MFE by the probabilistic loss functional. The key is to determine the \emph{data-dependent} function space of identifiability (FSOI), in which the quadratic loss functional has a unique minimizer. We consider two \emph{data-adaptive} $L^2$ spaces: one is unweighted with the Lebesgue measure, and the other is weighted with a data-dependent exploration measure. In each $L^2$ space, the second-order derivative of the loss functional defines a semi-positive integral operator, which acts as the operator of inversion. The FSOI is then the closure of this integral operator's eigenspace of nonzero eigenvalues. Furthermore, identifiability holds in the $L^2$ space if and only if the integral operator is strictly positive. However, the inverse problem is ill-posed due to the inversion of a compact operator.  Our results apply to both radial and non-radial interaction kernels.

Together with \cite{LLMTZ21,LiLu20}, this study completes a full characterization of the identifiability of kernels in interacting particle systems with either finitely or infinitely many particles. Notably, there are significant differences between these two settings. For systems with $N$ particles, the identifiability holds in the weighted $L^2$ space because of a coercivity condition with a constant $\frac{1}{N-2}$ (see \cite[Proposition 2.1]{LLMTZ21}), and the inverse problem is well-posed. In contrast, for the inverse problem of the MFE,  no coercivity holds in $L^2$ (in agreement with the above coercivity constant vanishes as $N\to\infty$), the identifiability barely holds in $L^2$, and the inverse problem is ill-posed.

The identifiability has important implications for computational practice. The ill-posedness implies that the normal matrix in regression becomes ill-conditioned as the dimension of the hypothesis space increases. Thus, regularization is necessary. The two ambient $L^2$ spaces provide natural norms for the Tikhonov regularization. We demonstrate numerically that the weighted $L^2$ norm is preferable over the unweighted $L^2$ space because it leads to more accurate regularized estimators in the context of singular value decomposition (SVD) analysis. 

Furthermore, the identifiability theory introduces adaptive RKHSs for regularization. They are different from the widely-used kernel regression \cite{FY03,rasmussen2003gaussian,MaestraHoffman22} or RKHS regularization \cite{cucker2007learning}, where the reproducing kernels are pre-selected. They invite further study on data-adaptive regularization strategies for ill-posed statistical learning and inverse problems \cite{LLA22}.

The exposition in our manuscript proceeds as follows. In Section \ref{sec:mainResults}, we define identifiability and introduce the main results. Section \ref{sec:radial} studies identifiability for radial kernels and Section \ref{sect:general} extends the results to general non-radial kernels. We discuss in Section \ref{sec:comput} the implications of identifiability to computational practice. Appendix  \ref{sec:appendix} provides a brief review of positive definite functions and reproducing kernel Hilbert spaces.  

We shall use the notations in Table \ref{tab:notation1}.
\begin{table}[htb]  \vspace{-2mm}
\centering
		\caption{  \text{Notations. }} \label{tab:notation1} \vspace{-2mm}
		\begin{tabular}{ l  l l }
		\toprule 
			   &  \quad  Radial kernel \quad \quad  & Non-radial kernel  \\  \hline
	 Interaction potential  & $\Psi(|x|)$ and $\psi = \Psi'$;  &  $  \Psi(x)$ \\
	 Interaction kernel      &  $K_\psi(x) = \psi(|x|)\frac{x}{|x|}$   & $K_\psi(x) = \grad \Psi(x)$  \\
	          Loss functional         & $\mE(\psi)$      & $\mE(K_\psi)$ \vspace{2mm} \\   \hline
	 	Density of exploration measure  & \multicolumn{2}{c}{$\rhoT$, $\mX =\mathrm{support}(\rhoT)$, in \eqref{eq:rhoavg}}  \\
	Function space of learning        & \multicolumn{2}{c}{$L^2(\mX)$ and $L^2_{\rhoT}$}  \vspace{2mm} \\ \hline 
	 Mercer kernel \& RKHS in $L^2(\mX)$ &      $\GbarT$ in \eqref{eq:GbarT_radial} and $\mH_G$   & $\FbarT$  in \eqref{eq:FbarT} and $\mH_F$\vspace{2mm} \\
	 Mercer kernel \& RKHS in $L^2_{\rhoT}$&      $\RbarT$ in \eqref{eq:RbarT} and $\mH_R$  & $\QbarT$  in  \eqref{eq:FbarT} and $\mH_Q$\\
	 			\bottomrule	  
		\end{tabular}  \vspace{-4mm}
\end{table}


\vspace{-2mm}\section{Main results}\label{sec:mainResults}

Consider the McKean-Vlasov mean-field equation (MFE) of interacting particles: 
\begin{equation}\label{eq:MFE}
	\begin{aligned}
  		\partial_t u &= \nu \Delta u + \divg [u  (K * u)], \ x\in\R^d,t\in [0,T] \\
 		u(x,t)&\geq 0, \, \int_{\R^d} u(x,t) dx = 1 \,,  
	\end{aligned}
\end{equation}
where  $K* u$ denotes the convolution 
\[
	(K* u)(x,t) = \int_{\R^d} K(y) u(x-y,t)dy. 
\]
Here $K= \grad \Phi :\R^d\to\R^d$ is called an \emph{interaction kernel} and $\Phi $ is called the interaction potential. In particular, if $\Phi $ is radial, denoting $\Phi (x) = \Phi (|x|)$ with an abuse of notation, we have
\begin{equation}\label{eq:K_Phi}
	K(x) = \grad ( \Phi (|x|)) =\phi (\abs{x}) \frac{x}{\abs{x}}, \text{ with } \phi (r) =\Phi '(r).  
\end{equation}
where $\phi $ is also called \emph{interaction kernel} for simplicity. 

The mean-field equation (also called aggregation-diffusion equation \cite{carrillo2019aggregation}) describes the macroscopic density for the systems of interacting particles when $N\to \infty$: 
\begin{equation}\label{eq:sod}
  \frac{d}{dt}{\bX_t^{i}} = - \frac{1}{N} \sum_{j= 1}^N K( \bX_t^{j} - \bX_t^{i})
  + \sqrt{2\nu} d\bB_t^i, \quad \text{for $i = 1, \ldots, N$},
\end{equation}
where $\bX_t^{i} \in \realR{d}$ represents the position of agent $i$ at time $t$. 
Denote by $\mu^{(N)}(dx,t)= \frac{1}{N}\sum_{i=1}^N \delta(\bX_t^{i}-x)$ the empirical measure of the particles. Under suitable conditions on $\Phi $,
 it is well-known that $\mu^{(N)}(dx,t) \rightarrow u(x,t)dx$ in relative entropy of the invariant measure as the number of particles $N \rightarrow \infty$ (see e.g., \cite{meleard1996_AsymptoticBehaviour,malrieu2003_ConvergenceEquilibriuma,carrillo2011_GlobalintimeWeak,jabin2017_MeanFielda}).

Our goal is to study the identifiability of the \emph{interaction kernel} $K$ or $\phi$ from data consisting of a solution of the MFE. 
Throughout the paper, we assume that the data $u$ is a bounded weak solution to the MFE:   
\begin{assumption}[Smoothness of data]\label{assumption_u}
The data $u$ is a bounded continuous weak solution to the MFE with bounded support, that is,  $\sup_{(x,t)\in\R^d\times [0,T]} |u(x,t)|\leq C_u$ for a positive constant $C_u$, and $\bigcup_{t\in [0,T]}\supp{u(\cdot,t)}$ is bounded, where $\supp{u(\cdot,t)} = \{x\in \R^d: u(x,t)>0 \}$.   
\end{assumption}
Such a solution exists when the interaction kernel is local Lipschitz with polynomial growth: 
\[
|K(x)- K(y) | \leq C (|x-y|\wedge 1) (1+|x|^m + |y|^m), \ \forall x,y\in \R^d
\]
for a constant $C>0$ and an integer $m\geq 1$. For further study on the forward problem of the MFE, we refer to \cite{sznitman1991_TopicsPropagation} for Lipschitz kernels,  \cite{meleard1996_AsymptoticBehaviour,malrieu2003_ConvergenceEquilibriuma} for uniform convex kernels and the existence of an equilibrium, and the references in \cite{carrillo2011_GlobalintimeWeak,jabin2017_MeanFielda,jabin2018_QuantitativeEstimates} for general (including singular) kernels. 
The assumptions on the solution being continuous with bounded support are technical, and we discuss possible extensions to measure-valued solutions with unbounded support in Remark \ref{rmk:assumption}.

In the rest of this section, we present the main results only for radial kernels, and similar results hold for non-radial kernels (see Section \ref{sect:general}). 

\vspace{-2mm}\subsection{A loss functional in nonparametric regression}
We consider nonparametric approaches in which one finds an estimator by minimizing a loss functional in a hypothesis space \cite{BFHM17,LangLu22,della2022nonparametric,yao2022mean}.  Importantly, noticing that the MFE in \eqref{eq:MFE} depends linearly on the kernel, we can estimate the kernel by nonparametric regression, in which we minimize a \emph{quadratic loss functionals} efficiently by solving a least squares problem.

We consider the probabilistic loss functional introduced in \cite{LangLu22}, 
 \begin{align}\label{eq:costFn}
  \mE(\psi)  = & \frac{1}{T}\int_0^T \int_{\R^d} [  \abs{K_\psi*u}^2 u  + 2  \partial_t u (\Psi(|\cdot|) *u)    - 2\nu u  (\Delta \Psi *u)  ] dxdt.
 \end{align}
 where $K_\psi(x) = \nabla \Psi(|x|) = \psi(|x|)\frac{x}{|x|}$. It is the expectation of the log-likelihood of the McKean-Vlasov stochastic differential equation (to be introduced in \eqref{eq:nSDE}).  It has two appealing features: (i) it is derivative-free (i.e., not using derivatives of the data $u$), which plays a key role in obtaining a robust estimator in \cite{LangLu22}; and (ii) it applies to high-dimensional systems because the integrals in $u$ can be written as expectations, which is important when $u$ is approximated by the empirical measure of the particles.

 To focus on the identification of the kernel, we present an oracle version of the loss functional (see Lemma \ref{lemma:oracle} for its derivation).   

\begin{definition}[Oracle loss functional]\label{def:costFn}
Let $u$ be a solution of the mean-field equation \eqref{eq:MFE} on [0,T], and denote $K_\psi(x) = \nabla \Psi(|x|) = \psi(|x|)\frac{x}{|x|}$, where 
$\Psi:\R^+ \to \R$ is a radial interaction potential with derivative $\psi(r) = \Psi'(r)$. 
We consider the loss functional  
\begin{equation}
\begin{aligned}\label{eq:costFn_theory}
 \quad  \mE(\psi)  & = \rkhs{\psi, \psi}  -2\rkhs{\psi, \phi_{true}}, 
  \end{aligned}
\end{equation}
where the bilinear form $\rkhs{\cdot,\cdot}$ is defined by \vspace{-2mm}
\begin{equation}\label{eq:bilinearform_0}
  \begin{aligned} 
  \quad   \rkhs{\varphi, \psi} : = \frac{1}{T}\int_0^T \int_{\R^d}   u \innerp{( K_\varphi*u)}{ ( K_{\psi}*u)} \ dx\ dt, \\
\end{aligned}
\end{equation}
assuming that the integrals are well-defined. 
\end{definition}

Since the loss functional is quadratic, its minimizer in a finite-dimensional hypothesis space $\mH$ is computed by least squares in practice (with the term involving the true kernel approximated using data).
\begin{remark}[Least squares estimator]\label{rmk:MLE_radial}
For any hypothesis space $\mH = \text{span}\crl{\phi_1, \dots, \phi_n}$ such that   
\begin{align*}
  A_n(i,j) = \rkhs{\phi_i, \phi_j}, \quad 
  b_n(i) = \rkhs{\phi_i, \phi_{true}}  
\end{align*}
are well-defined, a minimizer of the loss functional  $\mE$ in $\mH$ is given by least squares
\begin{equation*}
\widehat\phi_\mH = \sum_{i = 1}^n \widehat{c_i} \phi_i,  \text{ where } \widehat{c} = \argmin{c\in \R^n} \mE(c), \quad  \mE(c)= c^\top A_n c-2c^\top b_n.
\end{equation*}
In particular, when $A_n$ is invertible, we have $\widehat c=A_n^{-1}b_n$, and hence $\widehat \phi_\mH$ is the unique minimizer of $\mE$ in $\mH$. When $A_n$ is singular, $A_n^{-1}$ denotes the Moore--Penrose pseudo inverse. Furthermore, when $A_n$ is ill-conditioned and there are errors in the approximation of $b_n$ due to measurement noise or numerical error, regularization helps to avoid amplifying the errors in $b_n$ (see Section {\rm \ref{sec:comput}} for more discussions). 
\end{remark}
 In a parametric inference approach, the hypothesis space is determined by the parametric form of the kernel. In a nonparametric regression approach, one selects the optimal hypothesis space with proper smoothness and dimension. 

Two fundamental elements are crucial for both approaches: the function space of learning and identifiability. The function space of learning provides a proper metric on the accuracy of the estimator, and the identifiability reveals if the inverse problem is ill-posed and sheds insights on regularization. In the next two subsections, we first introduce data-adaptive function spaces of learning, and then define the identifiability.

\vspace{-2mm}\subsection{The data-adaptive function spaces of learning}\label{sec:FnSpace}
We consider two data-adaptive function spaces of learning: a weighted $L^2$ space with a data-based measure and an unweighted $L^2$ space on the support of the measure.  

We introduce first a \emph{exploration measure} to quantify the exploration of the kernel by data, because we can only learn the kernel in the region where the data explores. This measure originates from a probabilistic representation of the mean-field equation. Recall that Equation \eqref{eq:MFE} is the Fokker-Planck equation (also called the Kolmogorov forward equation) of the following McKean-Vlasov stochastic differential equation 
\begin{equation} \label{eq:nSDE}
\left\{
\begin{aligned}
d\barX_t =& -[\Ktrue*u](\barX_t,t)dt + \sqrt{2\nu}dB_t,\\
\mathcal{L}(\barX_t) = & u(x,t)dx,
\end{aligned}
\right.
\end{equation}
for all $t\geq 0$.  Here $\mathcal{L}(\barX_t)$ denotes the law of $\barX_t$, whose probability density is $u(\cdot,t)$. Let $(\barX_t',t\geq 0)$ be an independent copy of $(\barX_t,t\geq 0)$ and denote $r_t= \abs{\barX_t-\barX_t'}$. We write the convolution $K_\psi*u$ as 
\begin{equation}\label{eq:conv_expctn}
[K_\psi*u](\barX_t,t) = \E[K_\psi(\barX_t-\barX_t')\mid \barX_t]= \E[\psi(r_t) \frac{\barX_t-\barX_t'}{r_t} \mid \barX_t]. 
\end{equation}
This probabilistic representation of $K_\psi*u$ indicates that the independent variable of $\psi$ is explored by the process $\{|\barX_t-\barX_t'|, t\in[0,T]\}$ (or $\{\barX_t-\barX_t', t\in[0,T]\} $ for non-radial kernels). 
Let $\rhoTd$ denote the average of probability densities of the processes: 
\begin{equation} \label{eq:rhoavg} 
\begin{aligned}
\quad \rhoT(r) & = \frac{1}{T} \int_0^T \rho_t(r) dt= \frac{1}{T} \int_0^T\int_{\R^d}\int_{\mathbb{S}^{d}} r^{d-1}u(y-r\xi, t)u(y, t)d\xi dy dt,	
\end{aligned}
\end{equation}
where $\rho_t$ denotes the density  of  $|\barX_t'-\barX_t|$ (or $\barX_t'-\barX_t$ for the  general case) for each $t$. Under Assumption \ref{assumption_u}, the probability density function $\rhoT$ is $C^2$. We denote the support of $\rhoTd$ by $\mX$: 
\[
\mX := \supp{\rhoTd}.
\]
Note that $\mX$ is bounded since the set $\bigcup_{t\in [0,T]}\supp{u(\cdot,t)}$ is bounded.

Two data-adaptive function spaces emerge: $L^2(\mX,\rhoT(r)dr)$ (denoted by $L^2_{\rhoT}$ hereafter) and the unweighted space $L^2(\mX)$ with the Lebesgue measure on $\mX$. Both spaces are viable choices because the loss functional in \eqref{eq:costFn_theory} is well-defined in either of them (see Lemma \ref{lemma:costFn}). However, we will show by numerical examples that the latter extracts more information from data and leads to more accurate regularized estimators (see Section \ref{sec:comput}).

\vspace{-2mm}\subsection{Definition of identifiability}
We define identifiability as the uniqueness of the minimizer of a quadratic loss functional in a linear hypothesis space. This definition applies to general quadratic loss functionals. This study focuses on the loss functional in \eqref{eq:costFn_theory}. 

\begin{definition}[Identifiability]\label{def:identifiability} 
Given data consisting of a solution $(u(x,t), x\in \R^d, t\in [0,T])$ to the mean-field equation \eqref{eq:MFE} and a quadratic loss functional $\mE$, we say that the interaction kernel is \emph{identifiable} by $\mE$ in a linear subspace $\hypspace$ of $L^2(\mX)$ or $L^2_{\rhoT}$ if the true kernel is the unique minimizer of the loss functional in $\hypspace$. We call the largest such linear subspace the function space of identifiability (FSOI). 
\end{definition}

When $\mH$ is a finite-dimensional (e.g., in parametric inference), Remark \ref{rmk:MLE_radial} suggests that identifiability holds in $\mH$ if the normal matrix $A_n$ is invertible, in other words, $c^\top A_n c>0$ for all nonzero $c\in \R^n$. Similarly, when $\mH$ is infinite-dimensional, the identifiability is equivalent to the non-degeneracy of the bilinear form $\rkhs{\cdot,\cdot}$, as the following lemma shows. 
\begin{lemma}\label{lemma:Id_H}
Identifiability holds in a linear space $\mH$ for the loss functional $\mE$ in \eqref{eq:costFn} if the bilinear form $\rkhs{\cdot,\cdot}$ in \eqref{eq:bilinearform_0} is non-degenerate in $\mH$. That is, the true kernel is the unique minimizer of the loss functional $\mE$ in \eqref{eq:costFn_theory} in $\mH$ iff $\rkhs{\phi,\phi}>0$ for all $\phi\in \mH$ except $\phi=0$. 
\end{lemma}
\begin{proof} Denote the true kernel by $\phi_{true}$. Note that 
\begin{equation}\label{eq:mE_square}
\mE(\psi) = \rkhs{\psi-\phi_{true},\psi-\phi_{true}} - \rkhs{\phi_{true},\phi_{true}}. 
\end{equation}
Thus, $\phi_{true}\in \mH$ is the unique minimizer of $\mE$ iff $\rkhs{\phi,\phi}>0$ for all nonzero $\phi\in \mH$. 
\end{proof}

The bilinear form plays a key role in our study of identifiability. We can write it as (see \eqref{eq:bilinearform_derive} for its derivation) 
  \begin{align*}
  \rkhs{\varphi, \psi} & =\int_{\R^+} \int_{\R^+} \varphi(r)\psi(s)  \ \overline G_T(r,s)  \ dr\ ds =: \inp{\mL_\GbarT\phi,\psi}_{L^2(\mX)},  
\end{align*}
where the integral kernel $\GbarT$ is a Mercer kernel (see Lemma \ref{lem:G_Mercer}) given by 
\begin{equation}\label{eq:GbarT_radial}
 \begin{aligned}
\overline G_T(r,s) &=\frac{1}{T} \int_0^T 
\int_{\mathbb{S}^{d}} \int_{\mathbb{S}^{d}} \int_{\R^d} \innerp{\xi}{\eta} \,  (rs)^{d-1}  u(z-r\xi,t)u(z-s\eta, t)u(z,t) dzd\xi d\eta\, dt, 
\end{aligned}
\end{equation}
where $\mathbb{S}^{d}$ denotes the unit sphere in $\R^d$. Here $\mL_\GbarT$ denotes the integral operator with kernel $\GbarT$ (see its definition in \eqref{eq:LG}). It acts as the operator of inversion, and plays a key role in the connection between the function space of identifiability and the RKHS of $\GbarT$.

\vspace{-2mm}\subsection{Main results}
We characterize the data-dependent function spaces of identifiability (FSOI) of the loss functional \eqref{eq:costFn} in both $L^2(\mX)$ and $L^2_{\rhoT}$, and compare them in computational practice. 

\begin{itemize}
\item In $L^2(\mX)$, the FSOI is the closure of the RKHS $\mH_G$ with reproducing kernel $\GbarT$ in \eqref{eq:GbarT_radial}. The identifiability holds in any linear subspace of the FSOI. Importantly, the identifiability holds in $L^2(\mX)$ iff $\mH_G$ is dense in it, or equivalently, the integral operator $\mL_\GbarT$ in $L^2(\mX)$ with integral kernel $\GbarT$ is strictly positive (see Theorems \ref{thm:HG_id}-- \ref{thm:id_L2X}). 
\item In $L^2_{\rhoT}$, the same results hold with $\GbarT$ replaced by $\RbarT(r,s)= \frac{\GbarT(r,s)}{\rhoT(r)\rhoT(s)}$ 
(see Theorem \ref{thm:L2rho_id}). 
\item Similar identifiability results hold for non-radial kernels (see Section \ref{sect:general}).  
\end{itemize}

We point out that identifiability is weaker than well-posedness. Identifiability in a linear space $\mH$ only ensures that the loss functional has a unique minimizer in $\mH$. It does not ensure the well-posedness of the inverse problem unless the bilinear form satisfies a coercivity condition in $\mH$ (see Remark \ref{def_coercivty}). When $\mH$ is finite-dimensional, identifiability is equivalent to the invertibility of the normal matrix in regression (see Remark \ref{rmk:MLE_radial}), and identifiability implies well-posedness. However, when $\mH$ is infinite-dimensional, the inverse problem is ill-posed because the inverse of the integral operator $\mL_\GbarT$ is unbounded.

The identifiability study has important implications for computational practice. The identifiability theory implies that the regression matrix will become ill-conditioned as the dimension of the hypothesis space increases (see Theorem \ref{thm:ill-condition}). Thus, regularization becomes necessary. We compare two regularization norms, the norms of $L^2_{\rhoT}$ and $L^2(\mX)$, in the context of singular value decomposition (SVD) analysis and the truncated SVD regularization. Numerical tests suggest that the inversion in $L^2_{\rhoT}$ is less ill-conditioned and its regularization leads to more accurate estimators.


\vspace{-2mm}\section{Radial interaction kernels}\label{sec:radial}
Radial interaction kernels are of particular interest because of their simplicity and efficiency in representing symmetric interactions. We will first show that the loss functional is well-defined and prove its oracle version. 
Then, we discuss the identifiability in the ambient function spaces $L^2(\mX)$ and in $L^2_{\rhoT}$ in Section \ref{sec:IDL2}-\ref{sec:IDL2rho}, respectively. 

Throughout this section, we let $\psi \in L^2_{\rhoT} $ with $\rhoT$ in \eqref{eq:rhoavg}. We let $\Psi:\R^+\to\R$ be $\Psi(r) = \int_0^r \psi(s)ds$ and denote $K_\psi(x) =\psi(|x|)\frac{x}{|x|}$.

\vspace{-2mm}\subsection{The loss functional}\label{sect:costFn}
We show first that the oracle loss functional in \eqref{eq:costFn_theory} is well-defined and it is equivalent to the loss functional in practice. 
\begin{lemma}\label{lemma:costFn} For any $\varphi, \psi \in L^2_{\rhoT}$, the bilinear form $ \rkhs{\varphi, \psi}$ in \eqref{eq:bilinearform_0} satisfies  
\begin{align}\label{eq:bd_bilinearForm} 
 \rkhs{\varphi, \psi}\leq \|\psi\|_{L^2_{\rhoT}}\|\varphi\|_{L^2_{\rhoT}}. 
\end{align}
Also, for any $\psi\in L^2_{\rhoT}$, the radial loss functional  in \eqref{eq:costFn_theory} is bounded above by 
\begin{equation}\label{eq:costFn_upperBd}
\mE(\psi) \leq  \|\psi-\phi_{true}\|_{L^2_{\rhoT}}^2 - \rkhs{\phi_{true},\phi_{true}}. 
\end{equation}
\end{lemma}
\begin{proof} Recall that $u(\cdot,t)$ is the law of $\barX_t$ defined in \eqref{eq:nSDE}. 
 By definition in \eqref{eq:bilinearform_0},  we have
\begin{equation}\label{eq:bilinearExptn}
  \begin{aligned} 
 \rkhs{\varphi, \psi}
& =  \frac{1}{T}\int_0^T \int_{\R^d} \innerp{(K_{\varphi}*u)}{ (K_{\psi}*u)}   u(x,t)dx\ dt \\
& =  \frac{1}{T}\int_0^T \E[\innerp{ K_\varphi*u (\barX_t,t)} {K_\psi*u (\barX_t,t) }] dt. 
\end{aligned}
\end{equation}
Let $\barX_t' $ be an independent copy of $\barX_t$. The above integrand in time is controlled by 
\begin{align*}
& \E[\innerp{ K_\varphi*u (\barX_t,t)} {K_\psi*u (\barX_t,t) }] \\
\leq & \E[ | K_\varphi*u (\barX_t,t)|^2]^{1/2}   \E[ | K_\psi*u (\barX_t,t)|^2]^{1/2}  \quad \text{(Cauchy-Schwartz)} \\
 \leq  &  \E[ |\E [K_\varphi (\barX_t-\barX_t')| \barX_t]|^2] ^{1/2}\  \E[ |\E [K_\psi (\barX_t-\barX_t')| \barX_t]|^2]^{1/2}   \quad  \text{(by \eqref{eq:conv_expctn})}\\ 
 \leq & \E[ \E [|K_\varphi (\barX_t-\barX_t') |^2 | \barX_t]] ^{1/2}\, \E[ \E [|K_\psi (\barX_t-\barX_t') |^2 | \barX_t]] ^{1/2} \quad  \text{(by Jensen's inequality)}  \\
 = &    \E [|K_\varphi (\barX_t-\barX_t') |^2 ]^{1/2}\,  \E [|K_\psi (\barX_t-\barX_t') |^2 ]^{1/2} = \|\varphi\|_{L^2(\rho_t)} \|\psi\|_{L^2(\rho_t)}. 
\end{align*}
Then, we obtain \eqref{eq:bd_bilinearForm}. 

The upper bound in \eqref{eq:costFn_upperBd} follows from  \eqref{eq:bd_bilinearForm} and \eqref{eq:mE_square}. 
\end{proof}

\begin{lemma}\label{lemma:oracle}
The loss functional in \eqref{eq:costFn} can be written in the oracle version in \eqref{eq:costFn_theory} if $u$ is a solution of the MFE \eqref{eq:MFE} with kernel $\phi_{true}$.  
\end{lemma}

\begin{proof}
The proof follows from the MFE and integration by parts. More specifically, note that $u$ vanishes at the boundary of its support, we have $\int_{\R^d} u  (\Delta \Psi *u)  dx= \int_{\R^d} \Delta u  (\Psi *u) $ by integration by parts. Then,  the MFE $\partial_t u - \nu \Delta u =  \divg [u  (K_{\phi_{true}} * u)]$ implies that 
\begin{align*}
 & \frac{1}{T}\int_0^T \int_{\R^d} [  2  \partial_t u (\Psi(|\cdot|) *u)    - 2\nu u  (\Delta \Psi *u)  ] dxdt  \\
 =  &  \frac{1}{T}\int_0^T \int_{\R^d}  2  (\partial_t  u - \Delta u) (\Psi(|\cdot|) *u)  dxdt  \\
 =  & \frac{1}{T}\int_0^T \int_{\R^d}  2  \divg [u  (K_{\phi_{true}} * u)] (\Psi(|\cdot|) *u)  dxdt \\
 = &- \frac{1}{T}\int_0^T \int_{\R^d}   2  u  (K_{\phi_{true}} * u) \cdot (K_\psi *u)  dxdt  = -2 \rkhs{\psi, \phi_{true}}, 
 \end{align*}
where in the third equality, we used integration by parts along with the fact that $K_{\psi}(x) = \nabla\Psi(|\cdot|)$.   
\end{proof}

\vspace{-2mm}\subsection{Identifiability in the unweighted L2 space}\label{sec:IDL2}
Now we show that the interaction kernel is identifiable by the loss functional \eqref{eq:costFn} in the $L^2(\mX)$-closure of the RKHS $\mH_G$  with reproducing kernel $\GbarT$ defined in \eqref{eq:GbarT_radial}. The key element is the integral operator of this reproducing kernel: it connects the RKHS with the space $L^2(\mX)$ and allows for a spectral characterization of identifiability.

The reproducing kernel  $\GbarT$ emerges from the bilinear form in the loss functional. Specifically, by a change of variable to polar coordinates with $K_\varphi(y) =\varphi(|y|) \frac{y}{|y|} = \varphi(r) \xi$ by setting $r=|y|$ and $\xi =\frac{y}{|y|}\in\mathbb{S}^{d}$, we can write the bilinear form as
\begin{equation}\label{eq:bilinearform_derive}
  \begin{aligned} 
&  \rkhs{\varphi, \psi}
=  \frac{1}{T}\int_0^T \int_{\R^d} \innerp{(K_{\varphi}*u)}{ (K_{\psi}*u)}   u(x,t)dx\ dt   \\
= & \frac{1}{T} \int_0^T \int_{\R^d} \int_{\R^d} K_{\varphi}(y)  \cdot
K_{\psi}(z) \int_{\R^d} u(x-y, t) u(x-z, t)   u(x,t) dx dydz \ dt  \\  
= & \int_{0}^\infty \int_{0}^\infty \varphi(r)\psi(s)\GbarT(r,s) \ dr ds,  
\end{aligned}
\end{equation}
where the last equality follows from the definition of $\GbarT$ in \eqref{eq:GbarT_radial}, 
\[
\GbarT(r,s) =  \frac{1}{T} \int_0^T \int_{\mathbb{S}^{d}} \int_{\mathbb{S}^{d}} \int_{\R^d} \innerp{\xi}{\eta} \,  (rs)^{d-1}  u(x-r\xi,t)u(x-s\eta, t)u(x,t) dxd\xi d\eta  dt. 
\] 

\begin{lemma}\label{lem:G_Mercer}
Under Assumption {\rm\ref{assumption_u}}, the kernel $\GbarT$ is a Mercer kernel, i.e., it is symmetric, continuous and positive definite. 
\end{lemma}
\begin{proof} 	The symmetry is clear from its definition and the continuity follows from the continuity of $u$. 
 To show that it is positive definite (see Definition \ref{def_spd}), for any $\{r_1, \cdots, r_k\} \subset \R^+$ and $(c_1, \dots, c_k)\in\R^k$, we have
	\begin{align*}
		 & \sum_{i, j}c_i c_j \GbarT(r_i, r_j)   \\   = &\frac{1}{T} \int_0^T \int_{\R^d} \sum_{ij}c_ic_j (r_ir_j)^{d-1} 
		 \int_{\mathbb{S}^d } \int_{\mathbb{S}^d }   \xi\cdot\eta   u(x-r_i \xi,t)  u(x-r_j \eta,t)  d\xi d\eta  u(x,t) dxdt \\
		= & \frac{1}{T} \int_0^T \int_{\R^d}  \left| \sum_{i}c_i r_i^{d-1} \int_{\mathbb{S}^d }   \xi   u(x-r_i \xi,t) d\xi \right|^2 u(x,t) \,dxdt\geq0.
	\end{align*}
Thus, it is positive definite. 
\end{proof}

Since $\GbarT$ is a Mercer kernel, it determines an 
 RKHS $\mH_G$ with $\GbarT$ as reproducing kernel (see Appendix \ref{sec:appendix}). We show next that the identifiability holds on the closure of $\mH_G$, 
by studying the integral operator with kernel $\GbarT$: 
\begin{align}\label{eq:LG}
\mL_{\GbarT} f(r) = \int_{\R^+} \GbarT(r,s)f(s) ds.
\end{align}
Note that by definition, 
\begin{equation}\label{eq:bilinear_IntOp_G}
  	\rkhs{\varphi,\psi}= \inp{\varphi,\mL_{\GbarT} \psi}_{L^2(\mX)}. 
\end{equation}

We start with a lemma on the boundedness and integrability of $\GbarT$.
\begin{lemma}\label{lem:GbarT_integrability} 
Under Assumption \ref{assumption_u}, the kernel function $\GbarT$ in \eqref{eq:GbarT_radial} satisfies: 
\begin{itemize}
\item[(a)]  For all $r, s\in\R^+$, $\GbarT(r, s) \leq C_uC_d |\mX|^{d-1} \min\{\rhoTd(r), \rhoTd(s)\}$, where $|\mX|$ denotes the Lebesgue measure of $\mX$, and $C_d = |\mathbb{S}^{d}| = 2\pi^{d/2}\Gamma(\frac{n}{2}) $. 
\item[(b)] $\GbarT$ is in $L^2(\mX \times \mX)$ and $\frac{\GbarT}{\rhoT\otimes \rhoT} \in L^2(\rhoT\otimes \rhoT)$. 
\end{itemize}
\end{lemma}

\begin{proof} 
Recall that $\mX$ is the support of $\rhoT (r)= \frac{1}{T} \int_0^T\int_{\R^d}\int_{\mathbb{S}^{d}} r^{d-1}u(y-r\xi, t)u(y, t)d\xi dy dt$ defined in \eqref{eq:rhoavg}. Since $u$ has bounded support, so the set $\mX$ is bounded and $s\leq |\mX|$ for each $s\in \mX$. 
Then, by the uniform boundedness of $u$, we have $s^{d-1}\int_{\mathbb{S}^{d}} u(x-s\eta, t) d\eta \leq |\mX|^{d-1} C_u$ for any $s\in \R^+$ and $x\in \R^d$. Hence,
 \begin{align*}
\GbarT(r,s) &=  \frac{1}{T} \int_0^T \int_{\mathbb{S}^{d}} \int_{\mathbb{S}^{d}} \int_{\R^d} \innerp{\xi}{\eta} \,  (rs)^{d-1}  u(x-r\xi,t)u(x-s\eta, t)u(x,t) dxd\xi d\eta  dt \\
& \leq  |\mX|^{d-1} C_u  C_d  \frac{1}{T} \int_0^T \int_{\mathbb{S}^{d}} \int_{\R^d} r^{d-1}  u(x-r\xi,t)u(x,t) dxd\xi   dt \\
& = |\mX|^{d-1} C_u  C_d  \rhoT(r), 
\end{align*}
for any $s\in \mX$. Similarly, $\GbarT(r,s) \leq |\mX|^{d-1} C_uC_d \rhoT(s)$ for any $r$. Then, (a) follows. \\ 
\indent
For (b),  we obtain $\GbarT$ is in $L^2(\mX \times \mX)$ by applying (a):  
\begin{align*}
	&\int_{\R^+}\int_{\R^+} |\GbarT(r,s)|^2 dr ds  \leq  |\mX|^{2d-2} C_u^2  C_d^2 \int_{\R^+}\int_{\R^+}  \rhoT(r) \rhoT(s)drds = |\mX|^{2d-2} C_u^2  C_d^2. 
\end{align*}
Similarly,  we obtain $\RbarT(r,s) = \frac{\GbarT}{\rhoT\otimes \rhoT} \in L^2(\rhoT\otimes \rhoT)$ by applying (a) to get that  
that $\RbarT(r,s)^2 \rhoT(r)\rhoT(s)  = \frac{|\GbarT(r,s)|^2}{\rhoT(r)\rhoT(s)} \leq   |\mX|^{2d-2} C_u^2  C_d^2 $, and hence, 
\begin{align*}
	&\int_{\R^+}\int_{\R^+}\RbarT(r,s)^2 \rhoT(r)\rhoT(s) dr ds 
	  \leq   \int_{\mX}\int_{\mX} |\mX|^{2d-2} C_u^2  C_d^2  drds = |\mX|^{2d} C_u^2  C_d^2. 
\end{align*}
\end{proof}

By Lemma \ref{lem:GbarT_integrability} and Theorem \ref{thm:RKHS}, $\mL_\GbarT$ is a positive compact self-adjoint operator in $L^2(\mX)$, and it has countably many positive eigenvalues  $\{\lambda_i\}_{i = 1}^\infty$ with orthonormal eigenfunctions $\{\varphi_i\}_{i = 1}^\infty$ (note that the eigenfunctions of the eigenvalue $\lambda=0$ is excluded). In particular, $\{\sqrt{\lambda_i}\varphi_i\}_{i = 1}^\infty$ is an orthonormal basis of $\mH_G$. The following theorem follows directly. 
\begin{theorem}\label{thm:HG_id}
	The function space of identifiability by $\mE$ in \eqref{eq:costFn} in $L^2(\mX)$ is the $L^2(\mX)$-closure of $\mH_G$, the RKHS with reproducing kernel $\GbarT$ in \eqref{eq:GbarT_radial}.
\end{theorem}
\begin{proof}
By Lemma \ref{lemma:Id_H}, it suffices to show that $\rkhs{f,f}^2>0$ for any nonzero $f$ in the $L^2(\mX)$-closure of $\mH_G$. Since $\{\sqrt{\lambda_i}\varphi_i\}$ is an orthonormal basis of $\mH_G$, its $L^2(\mX)$-closure is the closure of the eigenspace $\mathrm{span}\{\varphi_i\}$ corresponding to nonzero eigenvalues. Thus, if $f = \sum_{i = 1}^\infty c_i  \varphi_i$ is nonzero, we have $\ang{f,f}_{L^2(\mX)} = \sum_{i = 1}^\infty c_i^2>0$, which ensures that $\rkhs{f,f}^2 = \sum_{i = 1}^\infty c_i^2 \lambda_i >0$. 
\end{proof}

The RKHS $\mH_G$ has the nice feature of being data-informed: its reproducing kernel $\GbarT$ depends solely on the data $(u(x,t), x\in\R^d, t\in [0,T])$. It provides a tool to investigate when the kernel is identifiable in $L^2(\mX)$.

\begin{theorem}[Identifiability in $L^2(\mX)$] \label{thm:id_L2X}
 For the loss functional $\mE$ in \eqref{eq:costFn}, the following statements are equivalent.
\begin{itemize}
\item[(a)] Identifiability holds in $L^2(\mX)$, i.e., $\rkhs{h,h} >0$ for any nonzero $ h\in L^2(\mX)$.
\item[(b)] $\mL_{\GbarT}$ is strictly positive. 
\item[(c)] $\mH_G$ is dense in $L^2(\mX)$.  
\end{itemize}
Moreover, for any $\phi =\sum_{i=1}^n c_i\varphi_i \in \mH = \mathspan\crl{\varphi_1, \dots, \varphi_n}$ with $\{\varphi_i\}$ being orthonormal eigenfunctions of $\mL_{\GbarT} $ corresponding to positive eigenvalues $\{\lambda_i\}$, we have 
\begin{align}\label{eq:3norms_G}
 \rkhs{\phi,\phi}  = \sum_{i = 1}^n c_i^2 \lambda_i ,  \quad   \|\phi\|^2_{\mH_G}  =\sum_{i = 1} ^n c_i^2 \lambda_i^{-1},\quad \|\phi\|^2_{L^2(\mX)}  =\sum_{i = 1} ^n c_i^2.
 \end{align}
 In particular, the bilinear form $\rkhs{\cdot, \cdot}$ satisfies the coercivity condition in $\mH$:
	\begin{align*}
		\rkhs{\phi,\phi} \geq \min \{ \sqrt{\lambda_i}\}_{i=1}^n  \|\phi\|_{L^2(\mX)}^2,\, \forall \phi\in \mH . 
	\end{align*}
\end{theorem}

\begin{proof} (a) $\Rightarrow$ (b). Suppose $\mL_\GbarT$ is not strictly positive, then there exists an eigenfunction $\phi \in L^2(\mX)$ corresponding to eigenvalue 0. But we would also have $\rkhs{\phi,\phi}= \ang{\mL_{\GbarT}\phi, \phi}_{L^2(\mX)} =0$.

	(b) $\Rightarrow$ (a). If $\GbarT$ is strictly positive, then $\crl{\varphi_i}_{i = 1}^\infty$ would be an orthonormal basis for $L^2(\mX)$ and all eigenvalues of $\mL_\GbarT$ are  positive. Take $\phi = \sum_{j = 1}^\infty c_j\varphi_j$. Then, $\rkhs{\phi,\phi} = 0$ implies 
	\begin{align*}
		\ang{\mL_\GbarT \phi, \phi}_{L^2(\mX)} = \langle{\sum_{j = 1}^\infty c_j\lambda_j  \varphi_j, \sum_{j = 1}^\infty c_j \varphi_j}\rangle_{L^2(\mX)} = \sum_{j = 1}^\infty c_j^2\lambda_j = 0 .
	\end{align*}
	Hence $\phi = 0$ in $L^2(\mX)$.

	(b) $\Leftrightarrow$ (c). Note that $\crl{\sqrt{\lambda_i} \varphi_i}_{i = 1}^\infty$ is a basis of $\mH_G$. Thus, $\mH_G$ is dense in $L^2(\mX)$ iff $\crl{\varphi_i}_{i = 1}^\infty$ is a basis of $L^2(\mX)$, i.e. $\mL_\GbarT$ is strictly positive.

	At last, for any $\phi = \sum_{j = 1}^n c_j\varphi_j\in \mH =  \mathspan\crl{\varphi_1, \dots, \varphi_n}$, we have 
	\begin{align*}
		\rkhs{\phi,\phi}^2 = \ang{\mL_\GbarT \phi, \phi}_{L^2(\mX)} = \sum_{j = 1}^n c_j^2\lambda_j \geq \min \{ \lambda_i\}_{i=1}^n  \|\phi\|^2_{L^2(\mX)}.
	\end{align*}
	Also, we have $\phi=\sum_{j = 1}^n c_j\lambda_j^{-1/2} \sqrt{\lambda_j} \varphi_j \in \mH_G$ with $\|\phi\|^2_{\mH_G}  =\sum_{i = 1} ^n c_i^2 \lambda_i^{-1}$;
\end{proof}

\vspace{-2mm}\subsection{Identifiability in the weighted L2 space}\label{sec:IDL2rho}
In this section, we study the identifiability in $L^2_{\rhoT}$ through the RKHS whose reproducing kernel is a weighted integral kernel. Since the results are mostly the same as those in $L^2(\mX)$, so we only briefly state the main results, then focus on discussing their relations. 

We define the following kernel $\RbarT$ on the set $\mX\times \mX$:
\begin{equation}\label{eq:RbarT}
	\RbarT(r,s) = \frac{\GbarT(r, \,s)}{\rhoTd(r)\rhoTd(s)}, \ \  (r,s) \in \mX\times \mX.
\end{equation}
The function $\RbarT$ is a positive definite kernel, since $\GbarT$ is by Lemma \ref{lem:G_Mercer}. Additionally, by Lemma \ref{lem:GbarT_integrability}, the kernel $\RbarT \in L^2(\rhoT\otimes\rhoT)$, so that  $\mL_\RbarT$. Thus, it defines a compact integral operator
\begin{align}\label{eq:operatorR}
\mL_\RbarT \varphi(s) = \int_{\R^+} \RbarT(r,s)\varphi(r) \rhoT(r) dr, 
\end{align} 
and it satisfies $  \rkhs{\varphi,\psi} = \inp{\varphi,\mL_{\RbarT} \psi}_{L^2_{\rhoT}} $. 

All the results for $L^2(\mX)$ extends to $L^2_{\rhoT}$. 
Importantly, $\mL_\RbarT$ is a positive compact self-adjoint operator in $L^2_{\rhoT}$,  and it has countably many positive eigenvalues  $\{\gamma_i\}_{i = 1}^\infty$ with orthonormal eigenfunctions $\{\psi_i\}_{i = 1}^\infty$. In particular, $\{\sqrt{\gamma_i}\psi_i\}_{i = 1}^\infty$ is an orthonormal basis of $\mH_R$. 
Similar to Theorem \ref{thm:HG_id}--Theorem \ref{thm:id_L2X}, the identifiability holds in $L^2_{\rhoT}$ if the RKHS $\mH_R$ is dense in it.  The following theorem summarizes these results. 
\begin{theorem}[Identifiability in $L^2_{\rhoT}$]\label{thm:L2rho_id}
	The function space of identifiability in $L^2_{\rhoT}$ is the $L^2_{\rhoT}$-closure of $\mH_R$, the RKHS with reproducing kernel $\RbarT$ in \eqref{eq:RbarT}. The following are equivalent.
	\begin{itemize}
	\item[(a)] Identifiability holds in $L^2_{\rhoT}$.
	\item[(b)] $\mL_{\RbarT}$ in \eqref{eq:operatorR} is strictly positive.
	\item[(c)] $\mH_R$ is dense in $L^2_{\rhoT}$.
	\end{itemize}
		Moreover, for any $\phi =\sum_{i=1}^n c_i\psi_i \in \mH = \mathspan\crl{\psi_1, \dots, \psi_n}$ with $\{\psi_i\}$ being orthonormal eigenfunctions of $\mL_{\RbarT} $ corresponding to eigenvalues $\{\gamma_i>0\}$, we have 
\begin{align}\label{eq:3norms_R}
 \rkhs{\phi,\phi}^2  = \sum_{i = 1}^n c_i^2 \gamma_i ,  \quad   \|\phi\|^2_{\mH_R}  =\sum_{i = 1} ^n c_i^2 \gamma_i^{-1},\quad \|\phi\|^2_{L^2_{\rhoT}}  =\sum_{i = 1} ^n c_i^2.
 \end{align}
 In particular, the bilinear form $\rkhs{\cdot, \cdot}$ satisfies the coercivity condition in $\mH$:
	\begin{align*}
		\rkhs{\phi,\phi} \geq \min \{ \sqrt{\gamma_i}\}_{i=1}^n  \|\phi\|_{L^2_{\rhoT}}^2,\, \forall \phi\in \mH . 
	\end{align*}
\end{theorem}

\begin{remark}[Relation to the coercivity condition of the bilinear form] \label{def_coercivty} 
Recall that  a bilinear form $\rkhs{\cdot,\cdot}$ is said to be coercive on a subspace $\mH\subset L^2_{\rhoT}$ if there exists a constant $c_\mH>0$ such that $\rkhs{\psi,\psi}\geq c_\mH \|\psi\|_{L^2_{\rhoT}}^2$ for all $\psi\in \mH$. Such a coercivity condition has been introduced on subspaces of $L^2_{\rhoT}$ in \cite{BFHM17,LZTM19,LMT21_JMLR,LMT21,LLMTZ21} for systems of finitely many particles. Theorem \ref{thm:L2rho_id} shows that, for any finite-dimensional hypothesis space $\mH = \mathspan\crl{ \psi_1, \dots,  \psi_n} $, the coercivity condition holds with $c_\mH=\min \{ \gamma_i \}_{i=1}^n$, but the coercivity constant vanishes as the dimension of $\mH$ increases to infinity. 
\end{remark}

Now we have two RHKSs, $\mH_G$ and $\mH_R$, whose closures in $L^2(\mX)$ and $L^2_{\rhoT}$ are the function spaces of identifiability. They are the images  $\mL_{\GbarT}^{1/2}(L^2(\mX))$ and $\mL_{\RbarT}^{1/2}(L^2_{\rhoT})$ (see Appendix \ref{sec:appendix}).  The following remarks discuss their relations. 
 
 \begin{remark}[The two integral operators] The integral operators $\mL_{\GbarT}$ and  $\mL_{\RbarT}$ are derived from the same bilinear form: for any $\varphi,\psi\in L^2(\mX)\subset L^2_{\rhoT}$,  we have 
 \[
 \rkhs{\varphi,\psi}  =  \inp{\mL_\GbarT \varphi, \psi}_{L^2(\mX)} = \inp{\mL_\RbarT \varphi, \psi}_{L^2(\rhoTd)}. 
 \]
 Since $L^2(\mX)$ is dense in $L^2_{\rhoT}$, the second equality implies that the null-space of and  $\mL_{\GbarT}$ is a subset of  $\mL_{\RbarT}$. 
 However, there is no correspondence between their eigenfunctions of nonzero eigenvalues. To see this, let $\varphi_i \in L^{2}(\mX)$ be an eigenfunction of $\mL_{\GbarT}$ with eigenvalue $\lambda_i>0$. Then, it follows from the second equality that 
 \[ \inp{\lambda_i \varphi_i, \psi}_{L^2(\mX)} =  \inp{\mL_\GbarT \varphi_i, \psi}_{L^2(\mX)} = \inp{\mL_\RbarT \varphi, \psi}_{L^2(\rhoTd)} = \inp{ \rhoT \mL_{\RbarT} \varphi_i,\psi}_{L^2(\mX)}
\]
  for any $\psi\in L^2(\mX)$. Thus, $\lambda_i\varphi_i  = \rhoT \mL_{\RbarT} \varphi_i$ in $L^2(\mX)$. Then, neither $\rhoT\varphi_i$ nor $\frac{\varphi_i}{\rhoT}$ is an eigenfunction of $\mL_{\RbarT}$.  
 \end{remark}

 \begin{remark}[Metrics on $\mH_G$ and $\mH_{R}$] We have three metrics on $\mH_G \subset L^2(\mX)$: the RKHS norm, the $L^2(\mX)$ norm and the norm induced by the bilinear form. By \eqref{eq:3norms_G}, these three metrics satisfy 
 \[
\sqrt{ \rkhs{\phi,\phi} }\leq \max_{i}\{ \sqrt{\lambda_i} \} \|\phi\|_{L^2(\mX)} \leq \max_{i}\{\lambda_i\} \|\phi\|_{\mH_G}
\]
 for any $\phi  \in \mH_G$. 
 Similarly, there are three metrics on $\mH_R\subset L^2_{\rhoT}$ satisfying, for any $\phi\in \mH_R$,  
 \[ 
 \sqrt{ \rkhs{\phi,\phi} } \leq \max_{i}\{ \sqrt{\gamma_i} \} \|\phi\|_{L^2_{\rhoT}} \leq \max_{i}\{\gamma_i\} \|\phi\|_{\mH_R} .
 \]   
 \end{remark}
\begin{remark}[Relaxing the assumption on data]\label{rmk:assumption}
It is possible to relax the technical assumptions that the solution $u$ is continuous with bounded support and consider measure-valued solutions or unbounded support. These assumptions are used to prove the integrability of the integral kernel $\GbarT$ in Lemmas {\rm \ref{lem:G_Mercer}-\ref{lem:GbarT_integrability}} so that the integrator $\mL_{\GbarT}$ is a bounded operator. But the $\mL_{\GbarT}$ can remain to be bounded when the data $u$ is measure-valued or has unbounded support. Furthermore, the weighted space $L^2(\rhoT)$ can be used to study the inference of singular kernels. We leave the more involving analysis as future work. 
\end{remark}


\vspace{-2mm}\section{Non-radial interaction kernels} \label{sect:general}
Non-radial interaction kernels are important because they provide more flexibility for modeling than radial kernels. We extend the identifiability analysis to non-radial kernels, using the same arguments as for the radial case. 

Throughout this section, we consider the non-radial vector-valued kernels  $K_\phi=  \nabla \Phi:\R^d\to \R^d$ being a gradient of an interaction potential $\Phi$, where the subscript $\phi$ emphasizes that the kernel is a gradient. We analyze the function space of identifiability by the loss functional in \eqref{eq:costFn}. The next theorem presents the main results.

\begin{theorem}[Identifiability of non-radial kernel]\label{thm:idL2rho}
Given data $( u(x,t): (x,t)\in\R^d\times [0,T])$ satisfying Assumption {\rm\ref{assumption_u}},  consider the estimation of the kernel $K_{true}:\R^d\to \R^d$ by minimizing the loss functional 
\begin{align}\label{eq:costFnK}
  \mE(K_\psi)   : = & \frac{1}{T}\int_0^T \int_{\R^d} \left[  \abs{K_\psi*u}^2 u 
+  2\partial_t u (\Psi *u)    + 2 \nu\grad u \cdot (K_\psi *u) \right ] dx\ dt, 
\end{align}
where $K_\psi(x) = \nabla \Psi(x)$ in either $L^2(\mX)$ or $L^2_{\rhoT}:=\{K:\R^d\to \R^d, \int_{\R^d}|K(x)|^2\rhoT(x)dx<\infty\}$, where $\mX = \mathrm{supp}(\rhoT)$ with $\rhoT$ is defined by 
 \begin{equation}\label{eq:rhoTd}
 \rhoTd(x) = \frac{1}{T} \int_0^T \rho_t(x) dt= \frac{1}{T} \int_0^T\int_{\R^d} u(y-x, t)u(y, t)dy dt. 
 \end{equation}
Let $F_t(x,y) : = \int_{\R^d} u(z-y, t) u(z-x, t)   u(z,t) dz$ and define $\FbarT$ and $\QbarT$ as 
\begin{align}\label{eq:FbarT}
 \FbarT(x,y) = \frac{1}{T} \int_0^T F_t(x,y) \ dt , \quad  \QbarT(x,y) = \frac{\FbarT(x, \,y)}{\rhoTd(x)\rhoTd(y)}, \ \ & (x,y) \in \mX \times \mX. 
\end{align}
Then, the function space of identifiability in $L^2_{\rhoT}$ by the loss functional $\mE$ is the $L^2_{\rhoT}$-closure of $\mH_Q$, the RKHS with $\QbarT$ as the reproducing kernel. Additionally, the following are equivalent: 
\begin{itemize}
\item[(a)] Identifiability holds in $L^2_{\rhoT}$.
\item[(b)] The operator $\mL_{\QbarT}: L^2(\rhoTd)\to L^2(\rhoTd)$ in \eqref{operatorLQ} is strictly positive: 
\begin{align}\label{operatorLQ}
	\mL_\QbarT f(x)= \int_{\R^d} \QbarT(y,x)f(y)\rhoTd(y)dy   
\end{align}
\item[(c)] $\mH_Q$ is dense in $L^2_{\rhoT}$.
\end{itemize}
Similarly, these claims hold in $L^2(\mX)$ by considering the $L^2(\mX)$-closure of $\mH_F$, the RKHS with $\FbarT$ as the reproducing kernel, and the corresponding integral operator $\mL_{\FbarT}$. 
\end{theorem}
\begin{proof}
The proof is mostly the same as the proofs of Theorems \ref{thm:HG_id}-- \ref{thm:id_L2X}. It consists of three steps.
\begin{itemize}
\item Show that $\QbarT$ and $\FbarT$ are square-integrable reproducing kernels, so their RKHSs are well-defined. Consequently, their integral operators $\mL_{\QbarT}$ and $\mL_{\FbarT}$ are semi-positive. 
\item Extend Lemma \ref{lemma:Id_H} to vector-valued functions by showing that loss functional $\mE$ has a unique minimizer in a linear space $\mH$ if and only if $ \rkhs{K,K}>0$ for any nonzero $K\in \mH$. Here $\rkhs{K_1,K_2}$ is a bilinear form for vector-valued functions $K_1, K_2 : \R^d \rightarrow \R^d$, defined by 
\begin{equation}\label{eq:bilinearF-d} 
\begin{aligned}
  & \rkhs{K_1,K_2}
	=  \frac{1}{T}\int_0^T \int_{\R^d} \innerp{(K_1*u)}{ (K_2*u)}  \  u(x,t)dx\ dt   \\
	= & \int_{\R^d} \int_{\R^d} K_1(y)  \cdot K_2(z)  \FbarT(y,z)  dydz =   \sum_{i=1}^d\int_{\R^d} \int_{\R^d} K_1^i(y) K_2^i(z)  \FbarT(y,z) dydz.  
\end{aligned}
\end{equation}
The extension is straightforward, because $\mE(K_\psi) = \rkhs{K_\psi, K_\psi} - 2\rkhs{K_\psi, K_{true}}$ if $K_{true}$ is the true kernel generating the data $u$.    
\item Show the spectral characterization of the FSOI and the equivalence between (a)-(c), which follow from the same proofs for Theorems \ref{thm:HG_id}-- \ref{thm:id_L2X}.  
\end{itemize}
Thus, we only need to prove that $\FbarT$ and $\QbarT$ are square-integrable reproducing kernels, which we do in Lemma \ref{lem:FbarT_noncompact} below. 
\end{proof}

\begin{lemma}\label{lem:FbarT_noncompact}
Under Assumption {\rm\ref{assumption_u}}, the functions $\FbarT,\QbarT:\R^d\to \R$ in \eqref{eq:FbarT} are symmetric, positive definite, and satisfy the following properties: 
\begin{itemize}
\item[(a)] For all $x, y\in\R^d$, $\FbarT(x, y) \leq C_u \rhoTd(x),$ where $C_u=\sup_{x\in \R^d,t\in [0,T]} u(x,t)$. 
\item[(b)] The function $\FbarT$ is in $L^2(\R^d\times \R^d)$, and $\QbarT$ is in $L^2_{\rhoT\otimes \rhoT}$. 
\end{itemize}
\end{lemma}

\begin{proof} 
Both functions are symmetry by definition. They are positive definite similar to the proof of Lemma \ref{lem:G_Mercer}. 

Part (a) follows from  \eqref{eq:FbarT} and that $C_u=\sup_{z\in \R^d,t\in [0,T]} u(z-x,t)$ for any $x$,  
\begin{align*}
        \FbarT(x,y) & = \frac{1}{T} \int_0^T  \int_{\R^d} u(z-y, t) u(z-x, t)   u(z,t) dz \, dt  \\
	& \leq C_u \frac{1}{T} \int_0^T\int_{\R^d} u(z-y, t)   u(z,t) dz \, dt = C_u \rhoTd(y), 
\end{align*}
where the last equality follows from the definition of $\rhoTd$.

For (b), note that by symmetry, we have $\FbarT(x,y) \leq C_u \rhoT(y)$ for any $x,y\in \R^d$. Then, 
	\begin{align*}
		\int_{\R^d} \int_{\R^d} (\FbarT(x,y))^2 dxdy \leq C_u^2 \int_{\R^d}\int_{\R^d}\rhoT(x)\rhoT(y)dxdy = C_u^2. 
	\end{align*}

To show that $\QbarT$ is square-integrable, we make use of the assumption that the data $u$ has bounded support, which implies that the support of $\rhoT$, denoted by $\mX= \mathrm{supp}(\rhoT)$,  is bounded. Hence, 
\[
\int_{\R^d}\int_{\R^d} \QbarT^2(x, y)\rhoT(x) \rhoT(y)dxdy = \int_{\mX} \int_{\mX} \frac{(\FbarT(x, y))^2}{\rhoT(x)\rhoT(y)} dxdy \leq C_u^2 |\mX|^2, 
\]
where the inequality follows from (a). 
\end{proof}

We note that the assumption on $u$ having a bounded support is sufficient but not necessary for $\QbarT\in L^2_{\rhoT\otimes \rhoT}$.  When the support of $u$ is unbounded, the function $\QbarT$ may not be square-integrable.  The following two examples show that $\QbarT$ is square-integrable when $u$ is the probability density function of a stationary Gaussian process, but it is not when $u$ is the density of a Cauchy distribution.


\begin{example}[Square-integrable $\QbarT$]\label{ex:Gaussian_steady_state}
	We show that  $\QbarT\in L^2_{\rhoT\otimes \rhoT}$ when $d=1$ and $u(x,0) = U(x) = \frac{1}{\sqrt{2\pi \nu} }e^{-\frac{x^2}{2\nu}}$. First, we show that $U(x)$ is a stationary solution to the mean-field equation \eqref{eq:MFE} (equivalently, $\mN(0, \nu)$ is an invariant density of the SDE \eqref{eq:nSDE}) with $K(x)=x$. In fact, noting that $K*U(x) = \int_\R (x-y)U(y)dy = x$, one can verify directly that $\nu U''(x) + [xU(x)]' = 0$ (similarly, the SDE \eqref{eq:nSDE} becomes the Ornstein-Uhlenbeck process $d\barX_t = - \barX_t dt + \sqrt{2\nu}dB_t$ and $U$ is its invariant density). Second, we compute $\FbarT$ and $\rhoTd$ directly from their definitions. Since $u(x,t) = U(x)$ for each $t$, by definition of $\FbarT$ in \eqref{eq:bilinearF-d}: 
	\begin{align*}
 		\FbarT(x,y) & = \int_{\R^d} U(z-x) U(z-y) U(z) dz  \\ 
		& = \,  \frac{1}{2\pi \nu} \int_{\R} \frac{1}{\sqrt{2\pi\nu}}e^{-\frac{1}{2\nu}( (z-x )^2 + (z-y)^2 + z^2)} dz =  \frac{\sqrt{3}}{2\pi \nu}e^{-\frac{1}{3\nu }(x^2 + y^2 - xy)}\ .
	\end{align*}
Since $\rhoTd$ is the density of $\barX_t-\barX_t'$ with $\barX_t'$ being an independent copy of $\barX_t$, which has the stationary density $U$, we have $\rhoTd(x) = \frac{1}{2 \sqrt{\pi\nu}}e^{-\frac{x^2}{4\nu}}$. Hence, the kernel $\QbarT$ is square-integrable due to the fast decay of $\FbarT$:
	\begin{align*}
	\|\QbarT\|_{L^2(\rhoTd \otimes \rhoTd)}^2 = \int_{\R}\int_\R \frac{\FbarT (x,y)^2}{\rhoTd(x)\rhoTd(y)} dx dy = \int_\R \int_\R \frac{3}{\pi\nu } e^{-\frac{1}{12 \nu^2} \edg{4(x-y)^2 + x^2 + y^2}} dx dy < \infty.
	\end{align*}
\end{example}

\begin{example}[Non-square-integrable $\QbarT$]\label{ex:Cauchy_steady_state}
We show that $\QbarT\notin L^2_{\rhoT\otimes \rhoT}$ when $d=1$, $\nu=1$ and $u(x,0) = U(x)= \frac{1}{\pi} \frac{1}{1 + x^2}$, which is the density of Cauchy distribution. Suppose that $K$ has a Fourier transform satisfying $\widehat K(\xi)  \widehat U (\xi)= \int_\R \frac{2 x}{1+x^2}e^{i\xi x}dx$, in other words, $[K*U](x) = \frac{2 x}{1+x^2}$. First, note that $U$ is a steady solution to \eqref{eq:MFE} because 
\[
\nu U''(x) =- \frac{d}{dx}\left[\frac{2 x}{1+x^2} U(x) \right]= -  \frac{d}{dx} \left [ U(x) [K*U](x)\right]. 
\]
Second,  direction computation (see Appendix \ref{sec:append_Example_computation} for the details) yields
	\begin{align*}
 		\FbarT(x,y) &= \int_{\R} U(z-x) U(z-y) U(z) dz = \frac{1}{\pi^3} \int_{\R} \frac{1}{1+(z-x)^2} \frac{1}{1+(z-y)^2} \frac{1}{1+z^2}dz  \\
&=    \frac{2}{\pi^2} \frac{(x^2 - xy + y^2 + 12)}{(x^2+4)(y^2+4) (x^2-2xy + y^2 + 4)}. 
	\end{align*}
	Meanwhile, since $\rhoTd$ is the density of $\barX_t-\barX_t'$ with $\barX_t'$ being an independent copy of $\barX_t$, which has the stationary density $U$, we have 
$ \rhoTd(x) = (U(\cdot)*U(-\cdot))(x) = \frac{2}{\pi}\frac{1}{(x^2 + 4)}$ (see Appendix \ref{sec:append_Example_computation} for the computation details). 
Lastly,  $\QbarT$ is not square integrable because 
	\begin{align*}
	\|\QbarT\|_{L^2(\rhoTd \otimes \rhoTd)}^2 = \int_{\R}\int_\R \frac{\FbarT (x,y)^2}{\rhoTd(x)\rhoTd(y)} dx dy = \int_\R \int_\R \frac{(x^2 - xy + y^2 + 12)}{2(x^2-2xy + y^2 + 4)} dx dy = \infty.
	\end{align*}
\end{example}


\vspace{-2mm}\section{Identifiability in computational practice}\label{sec:comput}

In this section, we discuss the implications of the identifiability theory for computational practice.  For simplicity, we consider only radial interaction kernels and $d=1$. We show that the regression matrix becomes ill-conditioned as the dimension of the hypothesis space increases (see Theorem \ref{thm:ill-condition}). Thus, regularization becomes necessary to avoid amplification of the numerical errors. We compare $L^2(\mX)$ and $L^2_{\rhoT}$ in the context of truncated singular value decomposition (SVD) regularization. Numerical tests in Section \ref{sec:regularization} suggest that the $L^2_{\rhoT}$ norm leads to more accurate regularized estimators, and a better-conditioned inversion (see Figure \ref{fig:estimattion_result}-- \ref{fig:Picard_cond}). 

\vspace{-2mm}\subsection{Nonparametric regression in practice}
In computational practice, the data is on discrete space mesh grids, and our goal is to find a minimizer of the loss functional by least squares as in Remark \ref{rmk:MLE_radial}. We review only those fundamental elements, and we refer to \cite{LangLu22} for more details. 

First, we select a set of data-adaptive basis functions in $L^2_{\rhoT}$ or $L^2(\mX)$ to avoid a singular normal matrix. The starting point is to approximate empirically the measure $\rhoT$ in \eqref{eq:rhoavg} from data and to obtain its support $\mX$. Let $\{r_i\}_{i = 0}^n$ be a uniform partition of $\mX$ and denote the width of each interval as $\Delta r$. They provide the knots for the B-spline basis functions of  $\mH$. Here we use piecewise constant basis functions to facilitate the rest discussions, that is, 
$  \mH = \mathrm{span}\{\phi_1, \dots, \phi_n\},\text{ with } \phi_i(r) = \mathbf{1}_{[r_{i-1},r_i]}(r)$.
 One may also use other partitions, for example, a partition with uniform probability for all intervals, as well as other basis functions, such as higher degree B-splines or weighted orthogonal polynomials. 
  
Second, as outlined in Remark \ref{rmk:MLE_radial}, we compute the normal matrix $A_n$ and vector $b_n$ from data.  Since $\phi_{true}$ is unknown, the vector $b_n$ is computed from data, following the loss functional in \eqref{eq:costFn}: 
  \begin{equation}  \label{eq:Abmat} 
 \begin{aligned}
 A_n(ij) &= \rkhs{\phi_i, \phi_j} =  \frac{1}{T}  \int_0^T \int_{\R^d}  (K_{\phi_i}*u)\cdot (K_{\phi_j}*u)   u(x,t)dx\ dt, \\
  b_n(i) &  = -\frac{1}{T}  \int_0^T \int_{\R^d} \left[ \partial_t u \, (\Phi_i *u)  +\nu \grad u \cdot ( K_{\phi_i} *u)\right] dx\ dt,    
 \end{aligned}
 \end{equation}
 where $\Phi_i(r) = \int_0^r \phi_i(s)ds$ is an anti-derivative of $\phi_i$.  
 The integrals in the entries of $A_n$ and $b_n$ are approximated from data by the Riemann sum.  
 
 Then, the minimizer of the loss functional in $\mH$ is solved from the linear equation $A_n \widehat c=b_n$.  
When the normal matrix $A_n$ is well-conditioned, we compute the minimizer by $\widehat{c}  = A_n^{-1}b_n$. When $A_n$ is ill-conditioned or singular, which happens often as $n$ increases, the (pseudo-)inverse of $A_n$ tends to amplify the numerical error in $b_n$. Thus, we need regularization  (see Section \ref{sec:regularization}).

\vspace{-2mm}\subsection{Identifiability and ill-conditioned normal matrix}
We show first that the eigenvalues of the integral operators are generalized eigenvalues of the normal matrix.  
\begin{theorem}\label{thm:ill-condition} 
Let $\mH = \mathrm{span}\{\phi_i\}_{i=1}^n \subset L^2(\mX)$, where the basis functions are linearly independent in $L^2(\mX)$ and $L^2_{\rhoT}$.  
Recall the normal matrix $A_n$ in \eqref{eq:Abmat}, the operators  $\mL_{\GbarT} : L^2(\mX) \to L^2(\mX) $ in \eqref{eq:LG} and $\mL_{\RbarT} : L^2_{\rhoT} \to L^2_{\rhoT} $ in \eqref{eq:operatorR}. The following statements hold true.
\begin{itemize}
\item[(a)] If $\mL_{\GbarT} \varphi = \lambda \varphi$ for some $\varphi =\sum_{i=1}^n c_i\phi_i \in \mH$, then $\lambda$ is a generalized eigenvalue of $A_n$:
\begin{equation}\label{eq:gEigenG}
A_n c= \lambda B_n^G c, \quad \text{  with } B_n^G  = (\inp{\phi_i, \phi_j}_{L^2(\mX)})_{1\leq i,j\leq n}.
\end{equation}
In particular, if $\{\phi_i\}$ are piecewise constants on intervals with length $\Delta r$ in a uniform partition of $\mX$, then, $\lambda\Delta r$ is an eigenvalue of $A_n$.  
\item[(b)] Similarly, if $\mL_{\RbarT} \varphi = \lambda \varphi$ for some $\varphi =\sum_{i=1}^n c_i\phi_i \in \mH$, then $\lambda$ is generalize eigenvalue of $A_n$:  
\begin{equation}\label{eq:gEigenR}
A_n c= \lambda B_n^R c, \quad \text{  with } B_n^R  = (\inp{\phi_i, \phi_j}_{L^2_{\rhoT}})_{1\leq i,j\leq n}.
\end{equation}
\end{itemize}
\end{theorem}
\begin{proof}
For Part (a), since $\mL_{\GbarT}\varphi = \lambda \varphi$ with $\varphi =\sum_{i=1}^n c_i\phi_i$,  we have
\[
\lambda (B_n^G c)_k = \inp{\lambda\varphi,\phi_k}_{L^2(\mX)} =  \inp{\mL_{\GbarT} \varphi, \phi_k}_{L^2(\mX)} =\sum_{i=1}^n c_i \inp{\mL_{\GbarT} \phi_i,\phi_k}_{L^2(\mX)} = (A_nc)_k,
\]
 where the last equality follows from \eqref{eq:bilinear_IntOp_G}. Thus, $\lambda$ is a generalized eigenvalue of $(A_n,B_n^G)$. Note that $B_n^G =\Delta r I_n$ if $\{\phi_i\}$ are piecewise constants on intervals with length $\Delta r$. Thus, by \eqref{eq:gEigenG}, $\lambda\Delta r$ is an eigenvalue of $A_n$.

Part (b) follows similarly. 
\end{proof}

We summarize the notations in these two generalized eigenvalue problems in Table \ref{tab:notation2}. 
\begin{table}[!htb] 
	\begin{center} 
		\caption{  Notation of variables in the eigenvalue problems on  $\mH$.} \label{tab:notation2}  \vspace{-2mm} 
		\begin{tabular}{ l  l l }
		\toprule 
			                        &  in $L^2(\mX)$  & in $L^2_{\rhoT}$ \\  \hline 
integral kernel and operator    & $\widehat{\GbarT}\approx \GbarT$, $\mL_{\widehat{\GbarT}} \approx \mL_{\GbarT}$ 
                                                & $\widehat{\RbarT} \approx \RbarT$, $\mL_{\widehat{\RbarT}}\approx \mL_{\RbarT}$\vspace{2mm} 
                                                \\
eigenfunction and eigenvalue  & $\mL_{\widehat{\GbarT}} \widehat{\varphi_k} =\widehat{\lambda}_k  \widehat{\varphi_k}$ 
	                                        & $\mL_{\widehat{\RbarT}} \widehat{\psi_k} =\widehat{\gamma}_k  \widehat{\psi_k}$\vspace{2mm}
 \\ 
	                                        eigenvector and eigenvalue  & $A_n \overrightarrow{\varphi_k} =\widehat{\lambda}_k  B_n^G \overrightarrow{\varphi_k}$ 
	                                        & $A_n \overrightarrow{\psi_k}= \widehat{\gamma}_k  B_n^R\overrightarrow{\psi_k}$ \\
			\bottomrule	
		\end{tabular}  \vspace{-4mm} 
	\end{center}
\end{table}		

\begin{remark}[Ill-conditioned  normal matrix] As the dimensions of $\mH = \mathrm{span}\{\phi_i\}_{i=1}^n$ increases, the normal matrix $A_n$ becomes ill-conditioned. This is because $A_n$ approximates the compact operators $\mL_{\GbarT}$ in $L^2(\mX)$ and $\mL_{\RbarT}$ in $L^2_{\rhoT}$, in the sense that 
\[
A_n(i,j) = \rkhs{\phi_i, \phi_j}  =  \inp{\mL_{\GbarT} \phi_i, \phi_j}_{L^2(\mX)} = \inp{\mL_{\RbarT} \phi_i, \phi_j}_{L^2_{\rhoT}}, \, \forall \phi_i, \phi_j\in \mH.
\]
Thus, Theorem {\rm \ref{thm:ill-condition}} indicates that, as $n$ increases, the generalized eigenvalues of $A_n$, with respect to $B_n^G$ and $B^R_n$, converge to those of $\mL_{\GbarT}$ and $\mL_{\RbarT}$, respectively. Then, the ratio $\lambda_{max}^{G, n}/\lambda_{min}^{G, n}$ increases to infinity, where we let $\lambda_{max}^{G, n}$ and $\lambda_{min}^{G, n}$ be the maximal and minimal generalized eigenvalues of $(A_n, B^G_n)$. 
Let $\lambda_{max}^{A_n}$ and $\lambda_{min}^{A_n}$ be the maximal and minimal eigenvalues of $A_n$, and similarly,  $\lambda_{max}^{B^G_n}$ and $\lambda_{min}^{B^G_n}$ for $B^G_n$.  Note that 
\begin{align*}
\lambda_{max}^{A_n} 
& = \max_{c\in \R^n}   \frac{c^\top A_n c}{c^\top B^G_n c} \frac{c^\top B^G_n c}{c^\top  c}
\geq  \max_{c\in \R^n}   \frac{c^\top A_n c}{c^\top B^G_n c}  \min_{c\in \R^n} \frac{c^\top B^G_n c}{c^\top  c}= \lambda_{max}^{G,n} \lambda_{min}^{B^G_n}, \\
\lambda_{min}^{A_n} &= 
\min_{c\in \R^n}   \frac{c^\top A_n c}{c^\top B^G_n c} \frac{c^\top B^G_n c}{c^\top  c}
\leq  \min_{c\in \R^n}   \frac{c^\top A_n c}{c^\top B^G_n c}  \max_{c\in \R^n} \frac{c^\top B^G_n c}{c^\top  c}=  \lambda_{min}^{G,n} \lambda_{max}^{B^G_n}.
  \end{align*}
Therefore, the conditional number of $A_n$ is bounded below as
\[
\frac{\lambda_{max}^{A_n}}{\lambda_{min}^{A_n}} \geq \frac{\lambda_{max}^{G, n} }{\lambda_{min}^{G,n}} \times  \frac{\lambda_{min}^{B^G_n}}{\lambda_{max}^{B^G_n}}. 
\]
Consequently, the matrix $A_n$ becomes increasing ill-conditioned as $\mH$ enlarges, since the ratio $\lambda_{min}^{B^G_n} / \lambda_{max}^{B^G_n}$ remains bounded for suitable basis functions. 
\end{remark}

\begin{remark}[Ill-posed inverse problem] The inverse problem is ill-posed in general: since $A_n$ becomes ill-conditioned as $n$ increases, a small perturbation in $b_n$ may lead to large errors in the estimator. More specifically, we are solving the inverse problem $\widehat \phi = \mL_{\RbarT}^{-1}( \mL_{\RbarT}\phi_{true})$ in $L^2_{\rhoT}$, where $\mL_{\RbarT}^{-1}$ is an unbounded operator. The normal matrix  $A_n$ approximates the operator $\mL_{\RbarT}$, and 
$b_n$ approximates $\mL_{\RbarT}\phi_{true}$. The error in $b_n$ in the eigenspace of small eigenvalues will be amplified by the inversion, leading to an ill-posed inverse problem. 
\end{remark}

\vspace{-2mm}\subsection{Truncated SVD regularization in the L2 spaces}\label{sec:regularization}
We compare the $L^2(\mX)$ and $L^2_{\rhoT}$ in the context of truncated Singular value decomposition (SVD) regularization. We show by numerical examples that the space $L^2_{\rhoT}$ leads to more accurate regularized estimators. 

\myparagraph{Truncated SVD regularization.} The truncated SVD regularization methods (see \cite{hansen1994_regularization_tools} and references therein) discard the smallest singular values of $A_n$ and solve the normal equation in the remaining eigenspace. To take into account the function spaces of learning, we present here a generalized version using generalized eigenvalues of $A_n$. More precisely, let  $B_n(i,j) = \inp{\phi_i,\phi_j}$ be a basis matrix, e.g., $B_n=B^G_n$ or $B^R_n$, which are extensions of $B_n= I_n$ in  \cite{hansen1994_regularization_tools}. Write 
\[A_n= \sum_{i=1}^n \sigma_i \bu_i \bu_i^\top,
\] where $\{\sigma_i\}$ are the decreasingly-ordered generalized eigenvalues of $(A_n,B_n)$ and $\{\bu_i\}\in \R^n$ are the corresponding $B_n$-orthonormal eigenvectors (i.e., $\bu_i^\top B_n \bu_j = \delta_{ij}$). The truncated SVD regularizer keeps only the $n_*$ largest eigenvalues above a proper threshold and leads to an estimator  
\begin{equation}\label{eq:estSVDreg}
\widehat \phi =  \sum_{j=1}^n \widehat c_j\phi_j, \quad \text{ where } \widehat c = \sum_{i=1}^{n_*} \frac{\bu_i^\top b}{\sigma_i} \bu_i. 
\end{equation}
This regularized estimator removes the error-prone contributions from $ \frac{\bu_i^\top b}{\sigma_i} \bu_i$ when $\sigma_i$ is small. Also, the estimator is regularized by expressing it as a linear combination of eigenfunctions corresponding to large eigenvalues, which have a resemblance to low-frequency trigonometric functions.



\myparagraph{Truncated SVD estimators in $L^2(\mX)$ and $L^2_{\rhoT}$.} To apply the truncated SVD regularization, we first compute the eigenvalues and eigenvectors corresponding to $L^2_{\rhoT}$ and $L^2(\mX)$. As suggested by Theorem \ref{thm:ill-condition}, they are from the generalized eigenvalue problems: 
 \begin{equation}\label{eq:reguSVD}
\begin{aligned}
 \text{Unweighted SVD }L^2(\mX) : \quad  	\mathbf\Phi \,D_G \,\mathbf\Phi^\top &= A_n, \text{ with } \mathbf\Phi^\top B^G_n \mathbf\Phi = I, \\
 \text{Weighted SVD } L^2_{\rhoT} : \quad  		\mathbf\Psi\, D_R \, \mathbf\Psi^\top &= A_n, \text{ with } \mathbf\Psi^\top  B^R_n \,\mathbf\Psi = I.
 \end{aligned}
\end{equation}
Here $D_G$ and $D_R$ are diagonal matrices consisting of the generalized eigenvalues of $(A_n, B^G_n)$ and $(A_n, B^R_n)$. 

We compare the truncated SVD estimators in $L^2(\mX)$ and $L^2_{\rhoT}$ for three examples: 
\begin{itemize}
\item cubic potential with $\phi(r) = 3r^2$; 
\item opinion dynamics with $\phi$ being piecewise linear; 
\item the attraction-repulsion potential with $\phi(r) = r-r^{-1.5 }$. 
\end{itemize}
These examples are studied in \cite{LangLu22},  and our numerical settings and simulations are the same as those in \cite[Section 4.1]{LangLu22} (except for simplicity we consider $v = 0.001$, $\Delta t= 0.01$ with $T=0.1$ and only $M=400$). 

We consider the basis functions being piecewise constants on a uniform partition of $\mX$. One can obtain better results by using spline basis functions with higher order regularity \cite{LangLu22}. Here the piecewise constant basis functions can highlight the regularity of the eigenfunctions in $L^2_{\rhoT}$ with $L^2(\mX)$, making it easier to compare $L^2_{\rhoT}$ with $L^2(\mX)$ in the truncated SVD regularization. The basis matrices become
\begin{align}\label{eq:P} 
B^G_n =\Delta r I_n, \quad B^R_n = \mathrm{diag}(\DrhoT(r_1), \dots, \DrhoT(r_n)) \Delta r  ,
\end{align}
where $ \{\DrhoT(r_i)\}$ is the average density on the interval $[r_{i-1}, r_i]$ , i.e. ${\DrhoT}(r_i) = \frac{1}{\Delta r}\int_{r_{i-1}}^{r_i} \rhoT(r) dr $. 
Note that we can represent $\rhoT$  by $\DrhoT = \sum_{i = 1}^n\DrhoT(r_i)\phi_i$.

\begin{figure}[htb]
	\captionsetup{width=1\linewidth}
 	\makebox[\textwidth][c]{
 		\subfigure[Cubic potential]{\includegraphics[scale=.32]{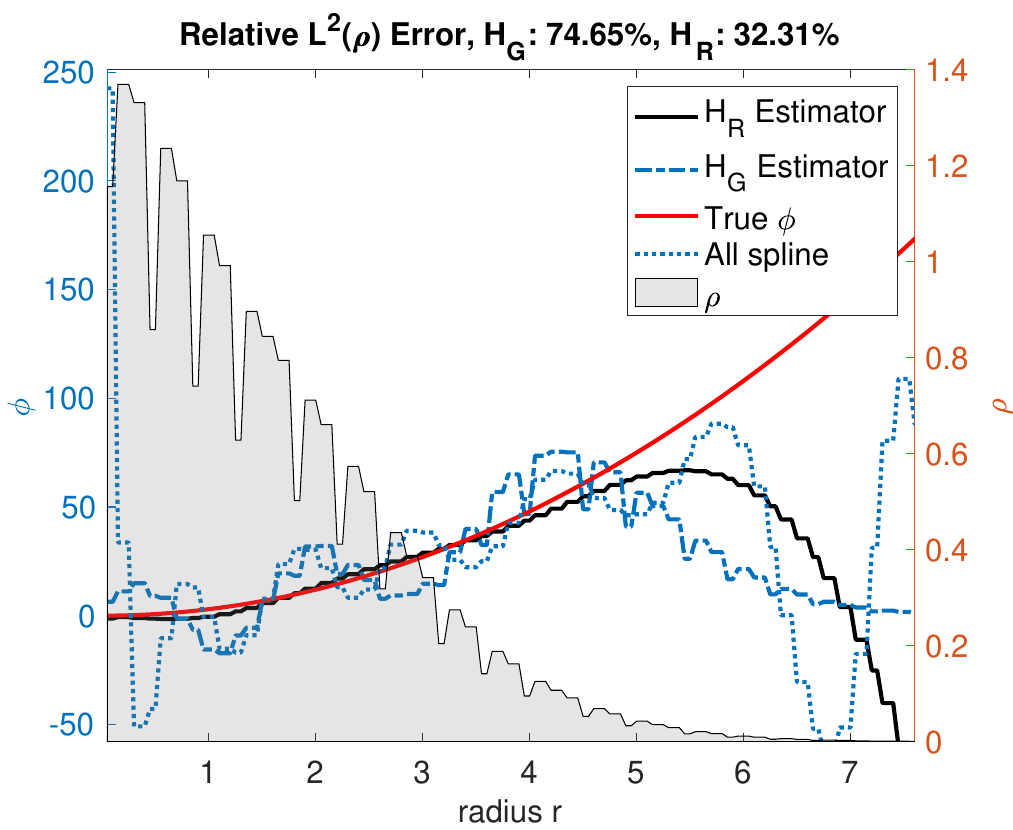} } \hspace{-2mm}
		\subfigure[Opinion dynamics]{\includegraphics[scale=.32]{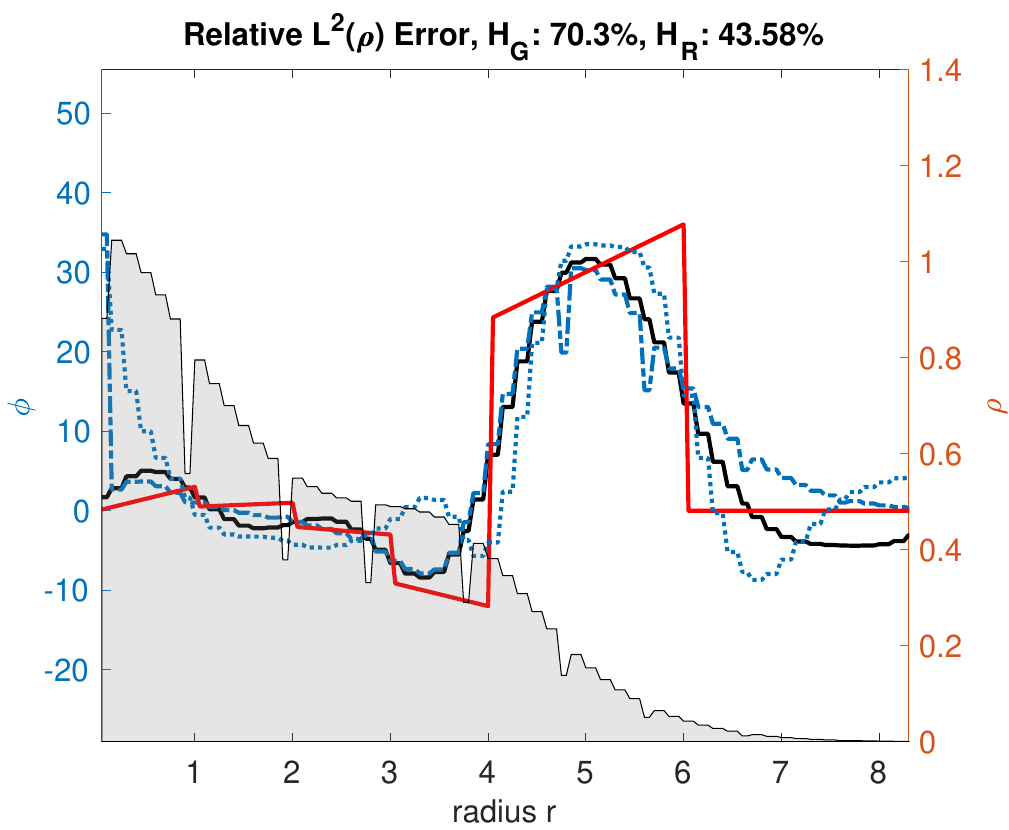}} \hspace{-2mm}
		\subfigure[Attraction-repulsion]{\includegraphics[scale=.32]{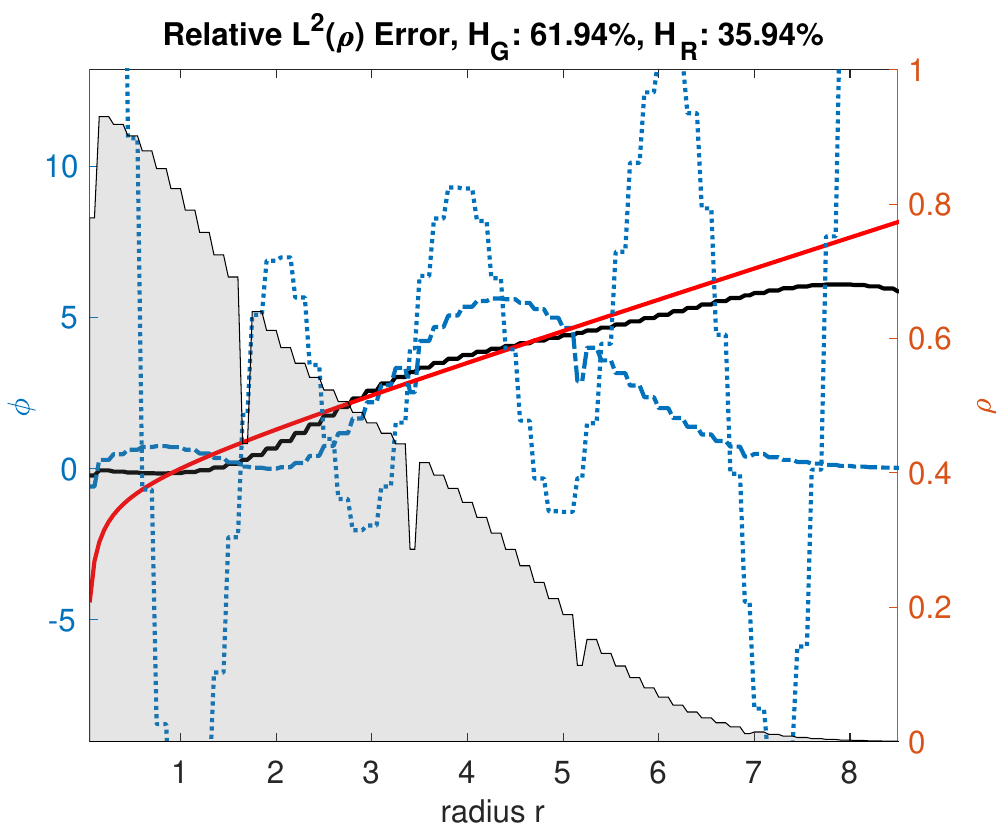}}
} \vspace{-4mm} 
\caption{Regularized estimators via truncated SVD for the three examples, superimposed with the exploration measure $\rhoT$. The weighted SVD (``$H_R$ Estimator'') has smaller errors than the unweighted SVD (``$H_G$ Estimator''), while both are significantly more accurate than the un-regularized estimator (``all spline'').
}\label{fig:estimattion_result}
 \vspace{-4mm} \
\end{figure}

\begin{figure}[htb] 
	\captionsetup{width=1\linewidth}
 	\makebox[\textwidth][c]{
 		\subfigure[Cubic potential]{\includegraphics[scale=.33]{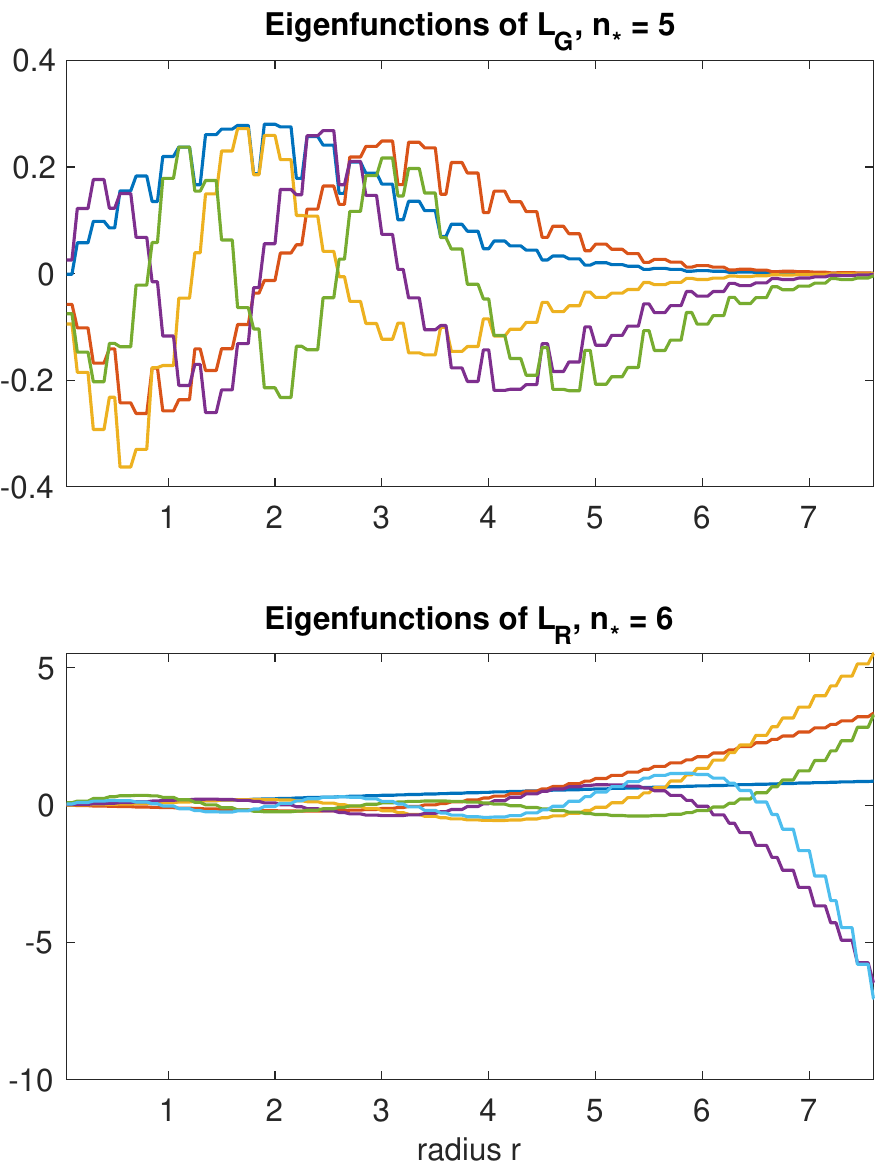} } \hspace{-2mm}
		\subfigure[Opinion dynamics]{\includegraphics[scale=.33]{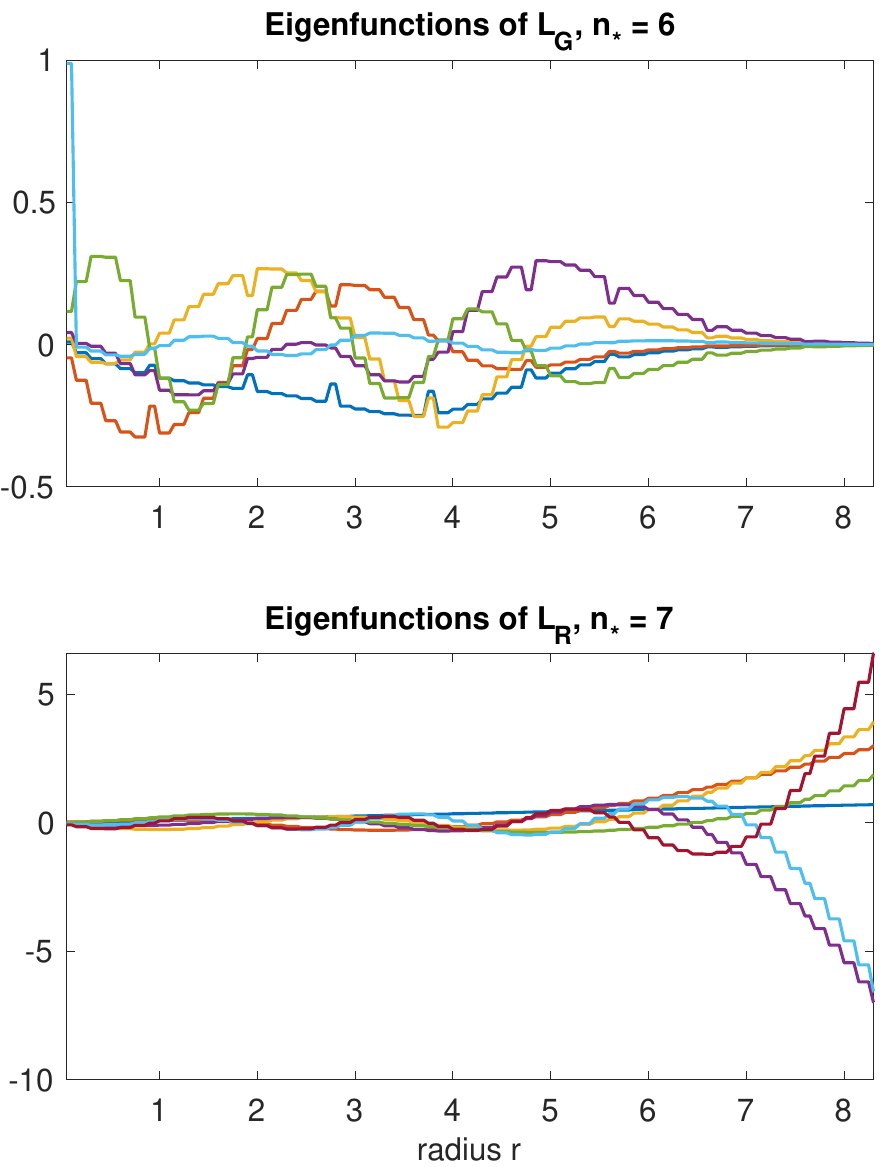}} \hspace{-2mm}
		\subfigure[Attraction-repulsion]{\includegraphics[scale=.33]{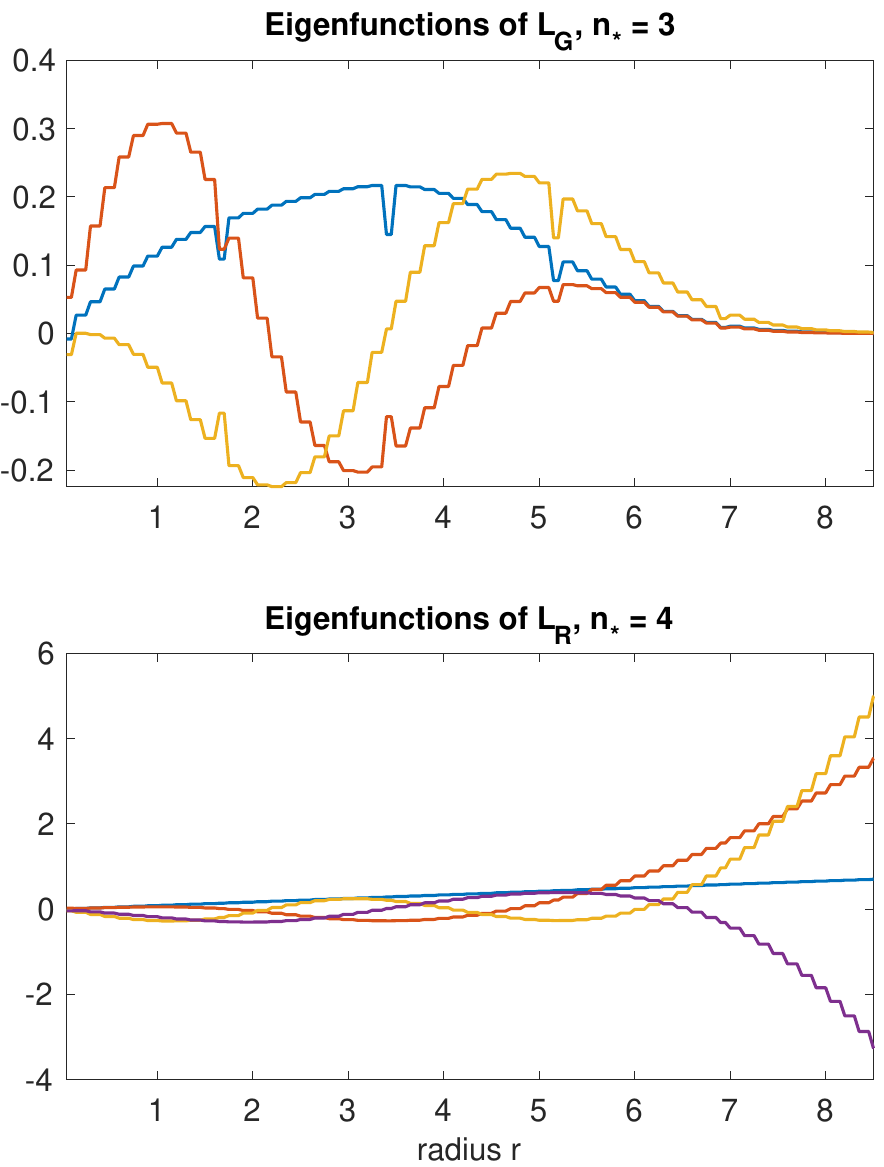}}
} \vspace{-4mm} 
\caption{Eigenfunctions in the estimation via weighted and unweighted SVD in Figure \ref{fig:estimattion_result}.  The weighted operator $\mL_R$ has smoother eigenfunctions than the unweighted operator $\mL_G$.
}
\label{fig:eigen_basis}
\end{figure}

Figure \ref{fig:estimattion_result} shows the regularized estimators via truncated SVD in $L^2(\mX)$ and $L^2_{\rhoT}$ for these examples. We pick a truncation level such that the sum of the largest $n_*$ singular values takes $99\%$ of the total summation of the singular values.  The corresponding eigenfunctions are presented in Figure \ref{fig:eigen_basis}. 
As can be seen, the weighted SVD leads to significantly more accurate estimators than the unweighted SVD.

\myparagraph{SVD analysis in $L^2(\mX)$ and $L^2_{\rhoT}$.} SVD analysis helps to understand the truncated SVD regularization. The truncated SVD regularization aims to remove the error-prone terms $ \frac{\bu_i^\top b}{\sigma_i} \bu_i$, particularly when the eigenvalue $\sigma_i$ is small. Thus, it is helpful to analyze the Picard ratio $ \frac{\bu_i^\top b}{\sigma_i}$ \cite{hansen1994_regularization_tools}. Clearly, when the ratio converges to zero, (called the discrete Picard condition),  the term $ \frac{\bu_i^\top b}{\sigma_i} \bu_i$ is error-immune; when the ratio increases largely, the inverse problem is ill-posed.

\begin{figure}[htb]
	\captionsetup{width=1\linewidth}
 	\makebox[\textwidth][c]{
 		\subfigure[Cubic potential]{\includegraphics[scale=.33]{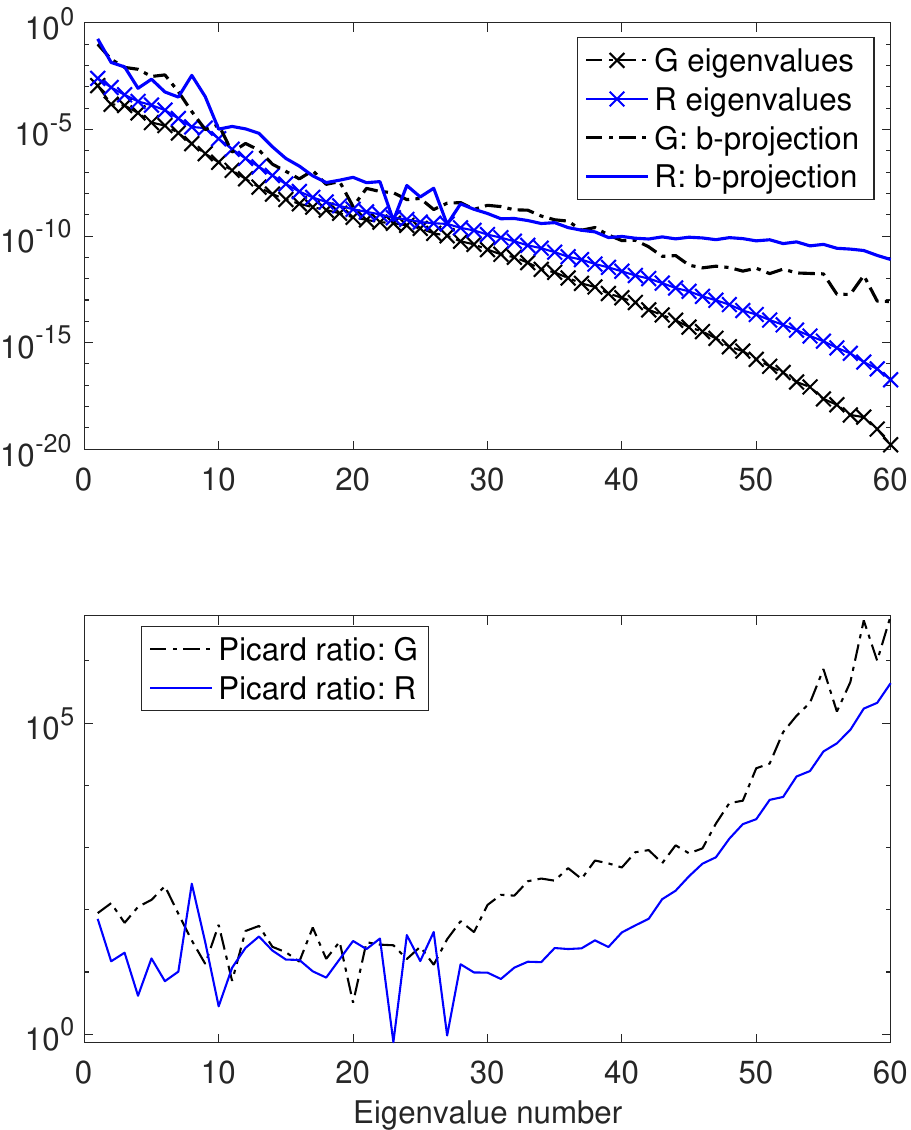} } \hspace{-2mm}
		\subfigure[Opinion dynamics]{\includegraphics[scale=.33]{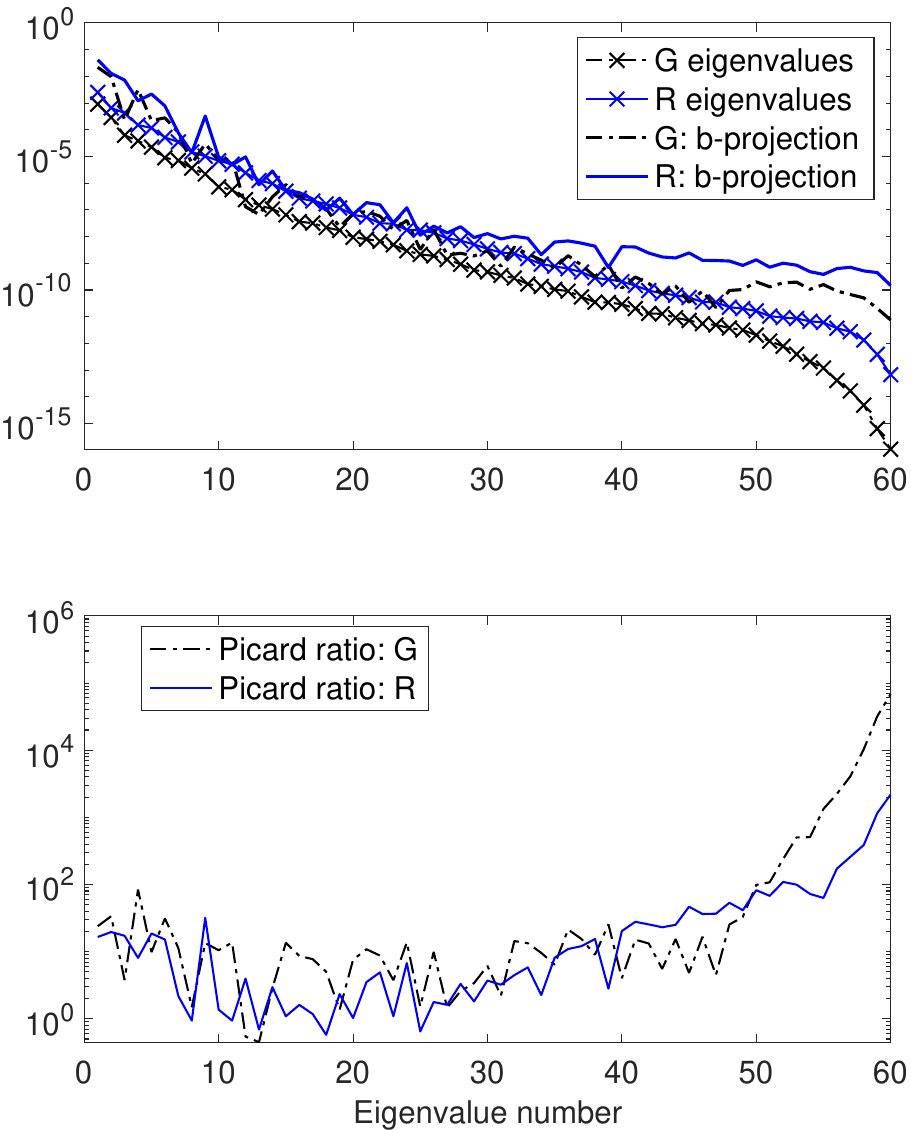}} \hspace{-2mm}
		\subfigure[Attraction-repulsion]{\includegraphics[scale=.33]{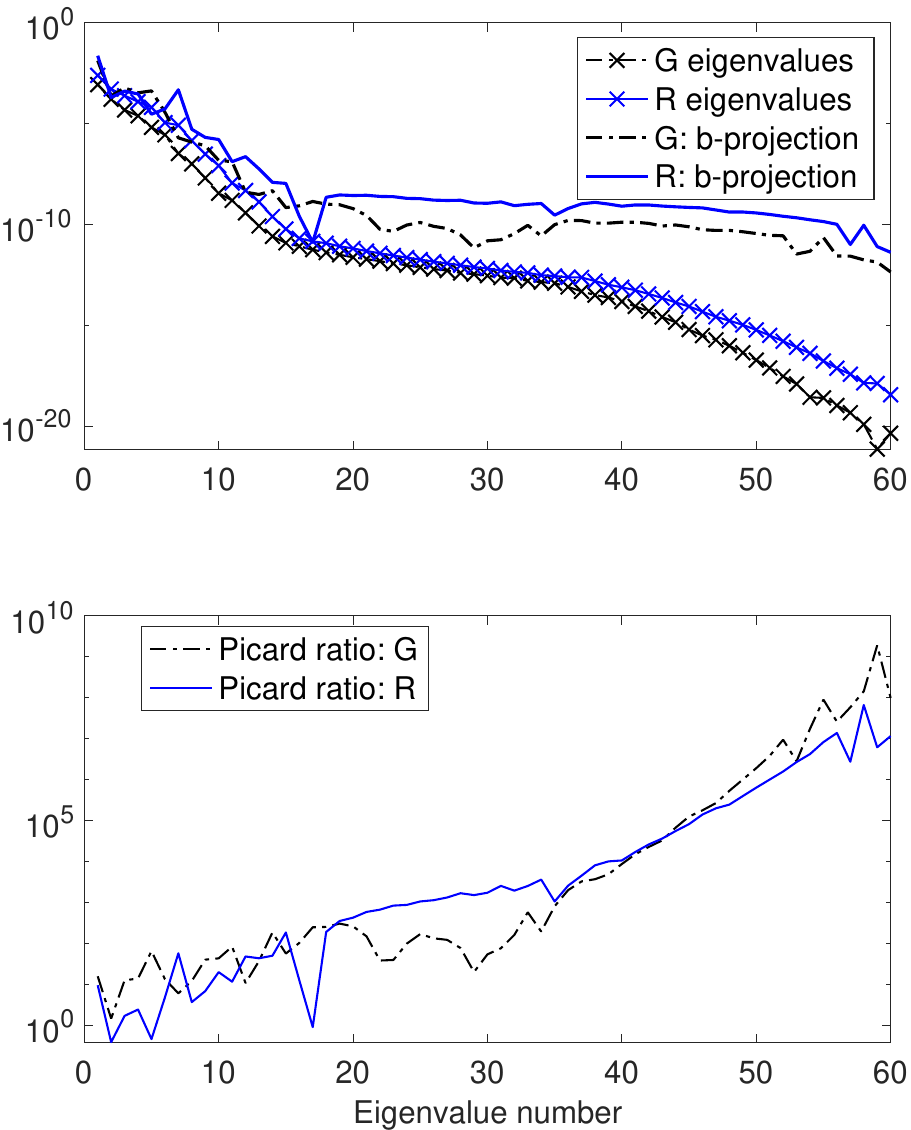}}
} \vspace{-4mm} 
\caption{SVD analysis of the regression in three examples.
Here $R$ represents the weighted SVD and $G$ represents the unweighed SVD.  In all three examples, the weighted SVD has larger eigenvalues than those of the unweighted SVD; and it has slightly smaller ratios $\frac{\bu_i^\top b}{\sigma_i}$. 
}\label{fig:Picard_cond} 
\end{figure}

Figure \ref{fig:Picard_cond} shows the singular values and the Picard ratios for the weighted and unweighted SVD \eqref{eq:reguSVD}. Here the unweighted SVD is denoted by $G$, and ``G: b-projection" refers to $|\mathbf u_i^\top b|$ with $\mathbf u_i$ being the columns of $\mathbf\Phi$. Similarly, $R$ denote the weighted SVD, and ``R: b-projection'' refers to $|\mathbf u_i^\top b|$ with $\mathbf u_i$ being the columns of $\mathbf\Psi$. In all these examples, the weighted SVD has larger eigenvalues than those of the unweighted SVD; and it has smaller Picard ratios. Thus, the weighted SVD leads to less ill-conditioned inversions and more accurate estimators. 

\begin{remark}[Tikhonov regularization with L-curve]\label{rmk:Lcurve_SVD}
The widely-used Tikhonov regularization \cite{hansen1994_regularization_tools} works well for this ill-posed inverse problem \cite{LangLu22}. 
It minimizes 
$	\mE_{\lambda}(\psi) = \mE(\psi) + \lambda \vertiii{\psi}^2,$
 where $\vertiii{\cdot}$ is a regularization norm. When the norm $\vertiii{\cdot}$ defines an inner product, it leads to a basis matrix $B_n$ for the basis functions $\{\phi_i\}_{i=1}^n$. The minimizer of $\mE_\lambda$ in the hypothesis space $\mH$ is  
$	\widehat c_\lambda = (A_n+\lambda B_n)^{-1} b_n.
$ 
 There are two factors in the method, the regularization norm and the hyper-parameter. Given a regularization norm $\vertiii{\cdot}$, an optimal parameter $\lambda$ aims to balance the decrement of the loss functional $\mE$ and the increment of the norm. Two groups of methods are successful. The L-curve method \cite{hansen_LcurveIts_a} selects $\lambda$ at where the largest curvature occurs on a parametric curve of ($\text{log}(\mE(\widehat c_\lambda)), \text{log}(\vertiii{\widehat c_\lambda}))$. The truncated SVD methods with the SVD analysis can also be used to select $\lambda$. However, the choice of regularization norm is problem-dependent and various norms have been explored, including the $H^1$-norm in \cite{LangLu22} and the RKHS norms (see \cite{LLA22} and the references therein).  
\end{remark}


\appendix
\vspace{-2mm}\section{Appendix}
\vspace{-2mm}\subsection{Review of RKHS and positive definite functions}\label{sec:appendix}
\myparagraph{Positive definite functions.}
We review the definitions and properties of positive definite kernels. The following is a real-variable version of the definition in \cite[p.67]{BCR84}.  

\begin{definition}[Positive definite function]\label{def_spd}
Let $X$ be a nonempty set. A function $G: X\times X\rightarrow \R$ is positive definite if and only if it is symmetric (i.e. $G(x,y)=G(y,x)$) and
$ \sum_{j,k=1}^{n}c_jc_kG(x_j,x_k)\geq 0 $
for all $n\in \mathbb{N}$, $\{x_1,\ldots,x_n\}\subset X$ and $\mathbf{c}=(c_1,\ldots,c_n) \in \R^n$. The function $\phi$ is strictly positive definite if the equality holds only when $\mathbf{c}=\mathbf{0} \in \R^n$. 
\end{definition}

 \begin{theorem}[Properties of positive definite kernels]\label{t52}
The following statements hold true. 
\begin{enumerate} \setlength\itemsep{0mm} 

\item[(a)]  Suppose that $k_1, k_2: X \times X \subset\mathbb{R}^d\times\mathbb{R}^d\to \mathbb{R}$ are positive definite kernels. Then, the product $k_1k_2$ is positive definite (\cite[p.69]{BCR84}). 



\item[(b)] Inner product $\langle u,v\rangle=\sum_{j=1}^du_jv_j$ is positive definite (\cite[p.73]{BCR84}). 

\item[(c)] $f(u)f(v)$ is positive definite for any function $f: X \to \mathbb{R}$ (\cite[p.69]{BCR84}).

\end{enumerate}
\end{theorem}

\myparagraph{RKHS and positive integral operators.} 
We review the definitions and properties of Mercer kernel, RKHS, and related integral operators on compact domains (see e.g., \cite{cucker2007learning}) and non-compact domains (see e.g., \cite{Sun2003Mercer}).  

Let $(X,d)$ be a metric space and $G:X\times X\to\R$ be continuous and symmetric. We say that $G$ is a Mercer kernel if it is positive definite (as in Definition \ref{def_spd}). The reproducing kernel Hilbert space (RKHS) $\mH_G$ associated with $G$ is defined to be closure of $\mathrm{span}\{G(x,\cdot):x\in X \}$ with the inner product 
\[
\langle f , g\rangle_{\mH_G} =  \sum_{i=1,j=1}^{n, m} c_i d_j G(x_i,y_j)
\]  
for any $f=\sum_{i=1}^n c_i G(x_i,\cdot)$ and $g=\sum_{j=1}^m d_j G(y_j,\cdot)$. It is the unique Hilbert space such that  $\mathrm{span}\{G(\cdot,y), y\in X\}$ is dense in $\mH_G$ and having reproducing kernel property in the sense that for all $f\in \mH_G$ and $x\in X$, $f(x) = \langle G(x,\cdot), f\rangle_{\mH_G}$ (see \cite[Theorem 2.9]{cucker2007learning}). 

By means of the Mercer Theorem, we can characterize the RKHS $\mH_G$ through the integral operator associated with the kernel. Let $\mu$ be a non-degenerate Borel measure on $(X,d)$ (that is, $\mu(U)>0$ for every open set $U\subset X$). Define the integral operator $\mL_G$ on  $L^2(X,\mu)$ by 
\[
\mL_Gf(x)  =\int_X G(x,y)f(y)d\mu(y). 
\] 
The RKHS has the operator characterization (see e.g., \cite[Section 4.4]{cucker2007learning} and \cite{Sun2003Mercer}).  
\begin{theorem}[Operator characterization of RKHS]\label{thm:RKHS}
Assume that the $G$ is a Mercer kernel and $G\in L^2(X\times X, \mu\otimes \mu)$. Then 
\begin{enumerate} \setlength\itemsep{0mm} 
\item $\mL_G$ is a compact positive self-adjoint operator. It has countably many positive eigenvalues $\{\lambda_i\}_{i=1}^\infty$ and corresponding orthonormal eigenfunctions $\{\phi_i\}_{i=1}^\infty$. 
\item $\{\sqrt{\lambda_i} \phi_i \}_{i=1}^\infty$ is an orthonormal basis of the RKHS $\mH_G$. 
\item The RKHS is the image of the square root of the integral operator, i.e., $\mH_G=\mL_G^{1/2} L^2(X,\mu)$. 
\end{enumerate}
\end{theorem}

\vspace{-2mm}\subsection{Computation details for Example \ref{ex:Cauchy_steady_state}}\label{sec:append_Example_computation}
We provide here the computation details in evaluating convolutions to obtain $\widebar{F}_T(x,y)$ and  $\widebar{\rho}_T(x,y)$ in Example \ref{ex:Cauchy_steady_state}. 

\myparagraph{1. Computation of $\widebar{F}_T(x,y)$.}
Recall that $U(x)= \frac{1}{\pi} \frac{1}{1 + x^2}$ and 
\[
\FbarT(x,y) = \int_{\R} U(z-x) U(z-y) U(z) dz = \frac{1}{\pi^3} \int_{\R} \frac{1}{1+(z-x)^2} \frac{1}{1+(z-y)^2} \frac{1}{1+z^2}dz. 
\]
Using a separation of rational functions, we can write 
\begin{equation}\label{eq:rational}
\frac{1}{1 + (z-x)^2}\frac{1}{1 + (z-y)^2}\frac{1}{1 + z^2} = \frac{Az + B}{1 + (z-x)^2} + \frac{Cz +D}{1 + (z-y)^2} + \frac{Ez +F}{1 + z^2},
\end{equation}
where each term can be integrated analytically. 
We first solve for the constants $A, B, C, D, E$ and $F$ from a system of linear equations that match the coefficients of the powers of $z^p$ with $p\in \{0,1,\ldots, 5\}$. For example, we have $A+C+E=0$ from the coefficient of $z^5$. 
Using a symbolic numerical solver, we obtain
\begin{align*}
	A& = \frac{-2(2x-y)}{x(x^2+4)((x-y)^3+4(x-y))}, \ \  B= \frac{5x^2-3xy-4}{x(x^2+4)((x-y)^3+4(x-y))},\\
	C& = \frac{2(x-2y)}{y(y^2+4)((x-y)^3+4(x-y))},  \ \  
	D = \frac{-5y^2+3xy+4}{y(y^2+4)((x-y)^3+4(x-y))},\\
	F& = \frac{xy-4}{xy(x^2+4)(y^2+4)}.
\end{align*}
Meanwhile, notice that each of the three terms in \eqref{eq:rational}  can be computed similarly and 
\begin{align} \label{eq:Rational_int}
\int_{\mathbb{R}}\frac{Az + B}{1 + (z-x)^2}dz &= \int_{\mathbb{R}}\frac{A(z-x) + (Ax+B)}{1 + (z-x)^2}dz  = A\int_{\mathbb{R}}\frac{t}{1 + t^2}dt+ (Ax+B)\int_{\mathbb{R}}\frac{1}{1 + t^2}dt.	
\end{align}
The terms with $\int_{\mathbb{R}}\frac{t}{1 + t^2}dt$ is canceled because $A+C+E=0$. Also 
$\int_{\mathbb{R}}\frac{1}{1 + t^2}dt = \pi.$ Hence, we have 
\begin{align*}
\widebar{F}_T(x,y) & = \frac{1}{\pi^2}[(Ax+B) + (Cy +D) + F]= \frac{2}{\pi^2} \frac{(x^2 - xy + y^2 + 12)}{(x^2+4)(y^2+4) (x^2-2xy + y^2 + 4)}.
\end{align*}

\myparagraph{2. Computation of $\widebar{\rho}_T(x,y)$.} The computation is similar to that of $\widebar{F}_T(x,y)$. 
Notice that, 
\begin{align*}
	\widebar{\rho}_T(x) = (U(\cdot)*U(-\cdot))(x) = \frac{1}{\pi^2}\int_{\mathbb{R}}\frac{1}{1 + (z-x)^2}\frac{1}{1 + z^2}dz.
\end{align*}
Thus, using a separation of rational functions 
$	\frac{1}{1 + (z-x)^2}\frac{1}{1 + z^2} = \frac{Az+B}{1 + (z-x)^2}\frac{Cz + D}{1 + z^2},
$ 
Eq.\eqref{eq:Rational_int}, and the fact that $A+C=0$ from the coefficient of $z^3$, we have 
\begin{align*}
	\widebar{\rho}_T(x) = \frac{1}{\pi}(Ax+B+D) = \frac{2}{\pi(x^2+4)}.	
\end{align*}

\ifarXiv
\textbf{Acknowledgements.} {The authors thank the two anonymous reviewers for their thoughtful and thorough comments.  
FL is grateful for supports from  NSF-1913243, FA9550-20-1-0288 and DE-SC0021361. FL would like to thank Mauro Maggioni and P-E Jabin for helpful discussions on mean-field equations and the inverse problem. }
\fi 
\bibliographystyle{myplain}\setlength{\itemsep}{0pt}
{\small

\begin{thebibliography}{10} 

\bibitem{baumgarten2019_GeneralConstitutive}
A.~S. Baumgarten and K. Kamrin.
\newblock A general constitutive model for dense, fine-particle suspensions
  validated in many geometries.
\newblock {\em Proc Natl Acad Sci USA}, 116(42):20828--20836, 2019.

\bibitem{bell2005_ParticlebasedSimulation}
N. Bell, Y. Yu, and P.~J. Mucha.
\newblock Particle-based simulation of granular materials.
\newblock In {\em Proceedings of the 2005 {{ACM SIGGRAPH}}/{{Eurographics}}
  Symposium on {{Computer}} Animation - {{SCA}} '05}, page~77, {Los Angeles,
  California}, 2005. {ACM Press}.

\bibitem{BCR84}
C. Berg, J.~P.~R. Christensen, and P. Ressel.
\newblock {\em Harmonic analysis on semigroups: theory of positive definite and
  related functions}, volume 100.
\newblock New York: Springer, 1984.

\bibitem{BFHM17}
M. Bongini, M. Fornasier, M. Hansen, and M. Maggioni.
\newblock Inferring interaction rules from observations of evolutive systems
  {I}: The variational approach.
\newblock {\em Mathematical Models and Methods in Applied Sciences},
  27(05):909--951, 2017.

\bibitem{carrillo2019aggregation}
J.~A. Carrillo, K. Craig, and Y. Yao.
\newblock Aggregation-diffusion equations: dynamics, asymptotics, and singular
  limits.
\newblock In {\em Active Particles, Volume 2}, pages 65--108. Springer, 2019.

\bibitem{carrillo2011_GlobalintimeWeak}
J.~A. Carrillo, M. DiFrancesco, A. Figalli, T. Laurent, and D. Slep{\v c}ev.
\newblock Global-in-time weak measure solutions and finite-time aggregation for
  nonlocal interaction equations.
\newblock {\em Duke Math. J.}, 156(2):229--271, 2011.

\bibitem{chen2021_MaximumLikelihood}
X. Chen.
\newblock Maximum likelihood estimation of potential energy in interacting
  particle systems from single-trajectory data.
\newblock {\em ArXiv200711048 Math Stat}, 2021.

\bibitem{cucker2007learning}
F. Cucker and D.~X. Zhou.
\newblock {\em Learning theory: an approximation theory viewpoint}, volume~24.
\newblock Cambridge University Press, 2007.

\bibitem{MaestraHoffman22}
L. Della~Maestra and M. Hoffmann.
\newblock The {LAN} property for {McKean-Vlasov} models in a mean-field regime,
  2022.

\bibitem{della2022nonparametric}
L. Della~Maestra and M. Hoffmann.
\newblock Nonparametric estimation for interacting particle systems:
  {McKean--Vlasov} models.
\newblock {\em Probability Theory and Related Fields}, 182(1):551--613, 2022.

\bibitem{FY03}
J. Fan and Q. Yao.
\newblock {\em Nonlinear Time Series: Nonparametric and Parametric Methods}.
\newblock Springer, New York, NY, 2003.

\bibitem{hansen1994_regularization_tools}
P.~C. Hansen.
\newblock {{REGULARIZATION TOOLS}}: {{A Matlab}} package for analysis and
  solution of discrete ill-posed problems.
\newblock {\em Numer Algor}, 6(1):1--35, 1994.

\bibitem{hansen_LcurveIts_a}
P.~C. Hansen.
\newblock The {L}-curve and its use in the numerical treatment of inverse
  problems.
\newblock In {\em in Computational Inverse Problems in Electrocardiology, ed.
  P. Johnston, Advances in Computational Bioengineering}, pages 119--142. WIT
  Press, 2000.

\bibitem{jabin2017_MeanFielda}
P.-E. Jabin and Z. Wang.
\newblock Mean {{Field Limit}} for {{Stochastic Particle Systems}}.
\newblock In N. Bellomo, P. Degond, and E. Tadmor, editors, {\em Active
  {{Particles}}, {{Volume}} 1}, pages 379--402. {Springer International
  Publishing}, {Cham}, 2017.

\bibitem{jabin2018_QuantitativeEstimates}
P.-E. Jabin and Z. Wang.
\newblock Quantitative estimates of propagation of chaos for stochastic systems
  with $w^{-1, \infty}$ kernels.
\newblock {\em Invent. math.}, 214(1):523--591, 2018.

\bibitem{kasonga1990_MaximumLikelihood}
R.~A. Kasonga.
\newblock Maximum {{Likelihood Theory}} for {{Large Interacting Systems}}.
\newblock {\em SIAM J. Appl. Math.}, 50(3):865--875, 1990.

\bibitem{LangLu22}
Q. Lang and F. Lu.
\newblock Learning interaction kernels in mean-field equations of first-order
  systems of interacting particles.
\newblock {\em SIAM Journal on Scientific Computing}, 44(1):A260--A285, 2022.

\bibitem{LiLu20}
Z. Li and F. Lu.
\newblock On the coercivity condition in the learning of interacting particle
  systems.
\newblock {\em arXiv preprint arXiv:2011.10480}, 2020.

\bibitem{LLMTZ21}
Z. Li, F. Lu, M. Maggioni, S. Tang, and C. Zhang.
\newblock On the identifiability of interaction functions in systems of
  interacting particles.
\newblock {\em Stochastic Processes and their Applications}, 132:135--163,
  2021.

\bibitem{LLA22}
F. Lu, Q. Lang, and Q. An.
\newblock {Data adaptive RKHS Tikhonov regularization for learning kernels in
  operators}.
\newblock {\em Proceedings of Mathematical and Scientific Machine Learning,
  PMLR 190:158-172}, 2022.

\bibitem{LMT21_JMLR}
F. Lu, M. Maggioni, and S. Tang.
\newblock Learning interaction kernels in heterogeneous systems of agents from
  multiple trajectories.
\newblock {\em Journal of Machine Learning Research}, 22(32):1--67, 2021.

\bibitem{LMT21}
F. Lu, M. Maggioni, and S. Tang.
\newblock Learning interaction kernels in stochastic systems of interacting
  particles from multiple trajectories.
\newblock {\em Foundations of Computational Mathematics}, pages 1--55, 2021.

\bibitem{LZTM19}
F. Lu, M. Zhong, S. Tang, and M. Maggioni.
\newblock Nonparametric inference of interaction laws in systems of agents from
  trajectory data.
\newblock {\em Proceedings of the National Academy of Sciences of the United
  States of America}, 116(29):14424--14433, 2019.

\bibitem{malrieu2003_ConvergenceEquilibriuma}
F. Malrieu.
\newblock Convergence to equilibrium for granular media equations and their
  {{Euler}} schemes.
\newblock {\em Ann. Appl. Probab.}, 13(2):540--560, 2003.

\bibitem{meleard1996_AsymptoticBehaviour}
S. M{\'e}l{\'e}ard.
\newblock {\em Asymptotic Behaviour of Some Interacting Particle Systems;
  {{McKean}}-{{Vlasov}} and {{Boltzmann}} Models}, volume 1627, pages 42--95.
\newblock {Springer Berlin Heidelberg}, {Berlin, Heidelberg}, 1996.

\bibitem{MT2014}
S. {Mostch} and E. {Tadmor}.
\newblock {Heterophilious Dynamics Enhances Consensus}.
\newblock {\em Siam Review}, 56(4):577 -- 621, 2014.

\bibitem{rasmussen2003gaussian}
C.~E. Rasmussen.
\newblock Gaussian processes in machine learning.
\newblock In {\em Summer school on machine learning}, pages 63--71. Springer,
  2003.

\bibitem{sharrock2021parameter}
L. Sharrock, N. Kantas, P. Parpas, and G.~A. Pavliotis.
\newblock Parameter estimation for the {McKean-Vlasov} stochastic differential
  equation.
\newblock {\em arXiv preprint arXiv:2106.13751}, 2021.

\bibitem{Sun2003Mercer}
H. Sun.
\newblock Mercer theorem for RKHS on noncompact sets.
\newblock {\em Journal of Complexity}, 21(3):337 -- 349, 2005.

\bibitem{sznitman1991_TopicsPropagation}
A.-S. Sznitman.
\newblock {\em Topics in Propagation of Chaos}, volume 1464, pages 165--251.
\newblock {Springer Berlin Heidelberg}, {Berlin, Heidelberg}, 1991.

\bibitem{VZ2012}
T. Vicsek and A. Zafeiris.
\newblock {Collective motion}.
\newblock {\em Physics Reports}, 517:71 -- 140, 2012.

\bibitem{yao2022mean}
R. Yao, X. Chen, and Y. Yang.
\newblock Mean-field nonparametric estimation of interacting particle systems.
\newblock {\em arXiv preprint arXiv:2205.07937}, 2022.

\end{thebibliography}

}

\end{document}